\documentclass{article}

     \PassOptionsToPackage{numbers, compress}{natbib}

     \usepackage[preprint]{neurips_2023}

\usepackage[utf8]{inputenc} 
\usepackage[T1]{fontenc}    
\usepackage{hyperref}       
\usepackage{url}            
\usepackage{booktabs}       
\usepackage{amsfonts}       
\usepackage{nicefrac}       
\usepackage{microtype}      
\usepackage{xcolor}         

\usepackage{amsmath}
\usepackage{amssymb}
\usepackage{amsthm}
\usepackage{algorithm}
\usepackage{algorithmic}
\usepackage{stmaryrd}
\usepackage{multirow}
\usepackage{changes}
\usepackage{comment}
\usetikzlibrary{patterns}
\usepackage{pgfplots}
\pgfplotsset{compat=1.18}
\usetikzlibrary{plotmarks}
\usepackage{booktabs}
\usepackage{subcaption}
\usetikzlibrary{shapes}

\theoremstyle{plain}
\newtheorem{theorem}{Theorem}[section]
\newtheorem{proposition}[theorem]{Proposition}
\newtheorem{lemma}[theorem]{Lemma}
\newtheorem{corollary}[theorem]{Corollary}
\theoremstyle{definition}
\newtheorem{definition}[theorem]{Definition}

\theoremstyle{remark}

\renewcommand{\vec}[1]{\boldsymbol{#1}}
\newcommand{\Prob}{\vec{p}}
\newcommand{\Dens}{\vec{p}}
\newcommand{\given}{\, | \,}
\newcommand*{\defeq}{\mathrel{\vcenter{\baselineskip0.5ex \lineskiplimit0pt
			\hbox{\footnotesize.}\hbox{\footnotesize.}}}%
	=}
\newcommand{\defi}{\defeq}

\newcommand{\fromto}{\longrightarrow}

\newcommand{\cX}{\mathcal{X}}

\newcommand{\cY}{\mathcal{Y}}
\newcommand{\cD}{\mathcal{D}}

\newcommand{\bX}{\mathbf{X}}
\newcommand{\bY}{\mathbf{Y}}

\newcommand{\bx}{\boldsymbol{x}}
\newcommand{\by}{\boldsymbol{y}}

\newcommand{\bh}{\boldsymbol{h}}

\newcommand{\GBNCs}{GBNCs}
\newcommand{\GBNC}{GBNC}

\DeclareMathOperator*{\argmax}{arg\,max}

\DeclareMathOperator*{\maximize}{Maximize}

\title{Probabilistic Multi-Dimensional Classification}

\author{%
  Vu-Linh Nguyen\footnote{These authors contributed equally} \\
  UTC, France\\
  \texttt{vu-linh.nguyen@hds.utc.fr} \\
 \And
 Yang Yang$^{\ast}$ \\
 KU Leuven, Belgium \\
 \texttt{yang.yang@kuleuven.be} \\
 \And
 Cassio de Campos \\
 TU/e, The Netherlands \\
 \texttt{c.decampos@tue.nl} \\
}

\begin{document}

\maketitle
\def\thefootnote{*}\footnotetext{These authors contributed equally to this work.}
\def\thefootnote{\arabic{footnote}}
\begin{abstract}
Multi-dimensional classification (MDC) can be employed in a range of applications where one needs to predict multiple class variables for each given instance. Many existing MDC methods suffer from at least one of inaccuracy, scalability, limited use to certain types of data, hardness of interpretation or lack of probabilistic (uncertainty) estimations. This paper is an attempt to address all these disadvantages simultaneously. We propose a formal framework for probabilistic MDC in which learning an optimal multi-dimensional classifier can be decomposed, without loss of generality, into learning a set of (smaller) single-variable multi-class probabilistic classifiers and a directed acyclic graph. Current and future developments of both probabilistic classification and graphical model learning can directly enhance our framework, which is flexible and provably optimal. A collection of experiments is conducted to highlight the usefulness of this MDC framework.
\end{abstract}

\section{Introduction}\label{sec:intro}
In (multi-class) classification, a predictive system makes use of a training data set (consisting of input-output pairs which specify individuals of a population) and a hypothesis space (consisting of the possible classifiers), and seeks for a classifier that optimizes its chance of making accurate predictions with respect to some given evaluation criterion (such as a loss function or an accuracy measure).
Numerous studies on classification have been devoted to learning probabilistic classifiers which predict, for each observation of the input space, a univariate probability distribution over the output space. The intention of probabilistic classification is not only to provide the end user with all necessary information about the optimal predictions of different loss functions \citep{elkan2001foundations,mortier2021efficient}, but also information about the uncertainty associated with the possible predictions. 

To overcome the assumption that the output space must be fully characterized by a single class variable, MDC has been proposed in which the output space is characterized by multiple class variables which can be correlated. MDC appears in important applications. An example of MDC is predicting subtypes/stages of diseases associated with each patient given his/her medical image and/or demographic information. Few other examples of MDC tasks are classification of biomedical text \citep{shatkay2008multi}, vehicle
classification \citep{jia2021decomposition} and beyond \citep{gil2021multi,jia2022multi}.

Existing multi-dimensional classifiers are non-probabilistic \citep{jia2021decomposition}, relatively inaccurate  \citep{jia2021decomposition}[Section II \& III], or unscalable \citep{gil2021multi}. To the best of our knowledge, no existing method specific for MDC is capable of directly handling mixed data, i.e., continuous and discrete features coexisting (without preprocessing or other external manipulations). Problem transformation methods \citep{jia2021decomposition} which transform the original MDC problem into either a huge multi-class classification (MCC) problem, for example using the class powerset (CP) classifier, or a set of independent MCC problems, for example the Binary relevance (BR) classifier, can be combined with deep multimodal learning \citep{kline2022multimodal,xu2021mufasa} to handle mixed data and other complex types of input. They suffer from the aforementioned issues and are arguably hard to interpret. The set of marginal probability distributions provided by BR can be associated to (infinitely) many joint distributions over the class variables\footnote{Thus, BR can be seen as a credal classifier and would be useful when targeting reliable set-valued predictions \citep{augustin2014introduction,jansen2022quantifying,troffaes2007decision}.} and does not inform much about the (true) joint distribution, while the joint distribution provided by CP contains an exponential number of masses and is not easily interpretable for end users. 

We present a framework to learn probabilistic multi-dimensional classifiers addressing those issues. This formal framework allows us to learn an optimal multi-dimensional classifier, without loss of generality/optimality, by decomposing the task into learning a set of probabilistic MCC models plus a directed acyclic graph (DAG). Notably, the framework inherits the interpretability of Bayesian networks (BNs) \citep{atienza2022hybrid,kitson2023survey,koller2009probabilistic}, which is a compact representation of quantitative and qualitative probabilistic relationships among class variables, and the scalability and flexibility of deep (multimodal) learning \citep{kline2022multimodal,lecun2015deep,xu2021mufasa}, i.e., handling complex types of data. Moreover, the probabilistic nature allows the framework, among other characteristics, to optimize different loss functions by only learning a single probabilistic model. We prove that the probabilistic model learned by this framework is universal and the learning procedure is globally optimal whenever MCC is universal and can be solved optimally too. We formalize the probabilistic MDC problem in Section~\ref{sec:probabilistic_MDC}, present formal results on the optimality of the framework in Section \ref{sec:GBNCs}, followed by a practical algorithm and properties of the learning framework in Sections~\ref{sec:Algorithmic_Solution} to~\ref{sec:Complexity_Learning_Problem}. Section~\ref{sec:Inference_Problem} discusses the inference task, and Section~\ref{sec:Experiments} further motivates the framework by presenting a collection of experiments indicating the advantages of the framework against existing MDC approaches. Section \ref{sec:Conclusion} concludes this paper. All formal results in this paper (propositions) are stated without proofs, which are deferred to Appendix \ref{sec:Proofs_of_Propositions_appendix}. 
Some technical details and experiments were also given in \citep{yang2022Generalized}.

\section{Probabilistic MDC} \label{sec:probabilistic_MDC}

Let $\mathbf{X} = \{X^1, \ldots, X^Q\}$ be a finite set of features, let $\cX:=\cX^1 \times \ldots \times \cX^Q$ denote an input space, and let $\mathbf{Y} = \{Y^1, \ldots, Y^K\}$ be a finite set of class variables. Let $\cY^k = \{y^{k,1}, \ldots, y^{k,M_k}\}$ be the set of $M_k$ possible outcomes for the $k^{\text{th}}$ class variable $Y^k$, $k \in [K] \defeq \{1, \ldots, K\}$. We define $\mathbf{Z} \defi \mathbf{Y} \cup \mathbf{X}$. We denote by $\mathbf{X}_d$ and $\mathbf{X}_c$ the discrete feature set and continuous feature set, respectively. We also define $\mathbf{Z}_d \defi \mathbf{Y} \cup \mathbf{X}_d$. For each instance $\bx \in \cX$, we say it is associated with a (vector)class $\by \in \cY =\cY^1 \times \ldots \times \cY^K$.  

We assume observations to be realizations of independently and identically distributed (i.i.d.) random variables generated according to a probability distribution on $\cX \times \cY$, i.e., an observation $\by=(y^1,\ldots, y^K)$ is the realization of a corresponding random vector $\bY = (Y^1, \ldots, Y^K)$. Let $\Dens(\mathbf{X}, \mathbf{Y})$ be a (mixed) joint density function. We denote by $\Prob(\mathbf{Y} \given \bx)$ the conditional joint distribution of $\bY$ given $\mathbf{X}=\bx$, whose probability mass function is given by
\begin{equation}\label{eq:jointConditionalProbabilities}
\Prob( \by \given \bx) \defi \frac{\Dens(\bx, \by)}{\sum_{\by' \in \cY} \Dens(\bx, \by')}  \, , \forall (\bx,\by) \in \cX \times \cY  \, .
\end{equation}
We assume the denominator to be non-zero whenever needed.
We denote by $\Prob(Y^k \given \bx)$, $k \in [K]$, the marginal distribution of $Y^k$, whose probability mass function is
\begin{align}\label{eq:marginalConditionalProbabilities}
\Prob( y^k \given \bx) &\defeq \sum_{\by\in\cY: Y^k = y^k} \Prob(\by \given \bx) \, , \forall y^k \in \cY^k \, .
\end{align}
Given training data in the form of a finite set of observations
$\cD = \big\{ (\bx_n,\by_n) \big\}_{n=1}^N  \subset \cX \times \cY$ 
drawn independently from a distribution, MDC aims to learn a predictive classifier model $\bh: \cX \fromto \cY$ assigning $\by \in \cY$ to each $\bx\in \cX$. The output of $\bh$ is a vector 
\begin{equation}\label{eq:h}
\hat{\by} \defi \bh(\bx) = \big(h^1(\bx), \ldots , h^K(\bx) \big) \in \cY \,  .
\end{equation}

In a probabilistic setting, a classification task can be viewed as a two-stage problem, in which a mapping $\bh: \cX \fromto \cY$ is not learned directly, but in a more indirect way. Roughly speaking, one can split a probabilistic classification into two tasks: learning a function $\Prob: \cX \fromto \Prob(\cY \vert \cX)$ (with abuse of notation) and constructing an efficient inference operator $o: \Prob(\cY \vert \cX) \fromto \cY$ (we will deal with $o$ in Section~\ref{sec:Inference_Problem}).

Motivated by the observations that discriminative models can perform better than generative models in many classification tasks \citep{bouchard2004tradeoff,carvalho2011discriminative,ng2001discriminative,ulusoy2006comparison}, and by the fact that in M-open cases~\citep{bernardo2000}, maximizing the (log) likelihood function may not converge to a best possible distribution as maximizing the conditional (log) likelihood function does \citep{roos2005discriminative}, we learn a multi-dimensional classifier encoding $\Prob$ which maximizes the conditional log likelihood (CLL) function $C(\Prob \given \cD)$:
\begin{equation}\label{eq:CLL}
    C(\Prob \given \cD) \defi \log \prod_{n=1}^N \Prob(\by_n \given \bx_n) \,.
\end{equation}
This idea has been mentioned before~\citep{benjumeda2018tractability}, but, to the best of our knowledge, it has been left open until now. Let $\mathcal{P}^0$ be a hypothesis space for $\Dens$. The learning problem can be defined as finding 
\begin{equation}\label{eq:Learning_problem}
   \Dens^* \in \argmax_{\Dens\in \mathcal{P}^0} C(\Dens \given \cD) \,.
\end{equation}
To avoid overfitting, the CLL function is often augmented by a regularization term. We will discuss it later.

\section{A Learning Framework} \label{sec:GBNCs}

The optimization problem \eqref{eq:Learning_problem} is very generic and its complexity highly depends on the given hypothesis space $\mathcal{P}^0$. We present reformulations of this problem that are more suitable to be optimized based on some assumptions about the hypothesis space. We proceed under the assumption that the features $X^q$, $q \in [Q]$, are always made available. This means we neither admit missing values at the training time \citep{nguyen2021racing} nor admit missing features at the prediction time \citep{saar2007handling}. This is not a limitation of the approach and missing data can be tackled using a variation of structure EM~\citep{friedman1998bayesian,rancoita2016}, but the discussion goes beyond the scope of this paper (see Appendix \ref{sec:Missing_data} for a quick discussion). 

Throughout, we assume the chain rule of probability \citep{koller2009probabilistic}[Section $2.1.3.4$] holds \footnote{An intensive study on the conditions under which the chain rule of probability is (in)valid is beyond the scope of this paper.}. Using the concept of conditional independence, we can assume without loss of generality that any $\Dens(\mathbf{X},\mathbf{Y})$ can be fully encoded by a DAG $G$ and a parameter set $\theta$ inducing the factorization
\begin{equation}\label{eq:our_assumption}
     \Dens^G_\theta(\bx, \by) = \prod_{X \in \mathbf{X}_c}\Dens_\theta(x \given \pi_x)\prod_{Z \in \mathbf{Z}_d}\Prob_\theta(z \given \pi_z) \,,  
\end{equation}
where $\pi_x$ and $\pi_z$ are (with abuse of notation) called configurations (compatible with $(\bx,\by)$) of the parent sets $\Delta_G^X$ and $\Delta_G^Z$ (for easiness, we assume that discrete parts of configurations are dictionaries with pairs (variable, value), and continuous parts are given via the appropriate functionals). The complexity of this factorization depends on $G$.

Therefore, the hypothesis space of any probabilistic MDC can be defined as $\mathcal{P} \defi \mathcal{G}\times \Theta$, where $\mathcal{G}$ and $\Theta$ are respectively the set of possible DAGs and the set of possible parameter sets, and the problem \eqref{eq:Learning_problem} becomes 
\begin{equation}\label{eq:Learning_problem_BN}
\Dens^{G^*}_{\theta^*}: ~ (G^*, \theta^*)  \in \argmax_{(G, \theta) \in \mathcal{P}} C(\Dens^G_\theta \given \cD)\,.
\end{equation} 

A learning procedure is optimal if it can find an optimal pair $(G^*,\theta^*)$. Parameter learning is optimally solved if we can find $\theta^*$ in~\eqref{eq:Learning_problem_BN} for a given $G\in\mathcal{G}$. In the following, we show that the factorization in~\eqref{eq:our_assumption} can lead to a great simplification of the learning problem \eqref{eq:Learning_problem_BN}.

\begin{proposition}\label{pro:Learning_problem_BND}
Assume the parameter learning problem is optimally solved. We have
\begin{equation}\label{eq:Learning_problem_BND}
\max_{\Prob\in\mathcal{P}^0} C(\Prob \given \cD) =\!\!\! \max_{(G, \theta) \in \mathcal{P}} C(\Prob^G_\theta \given \cD) =\!\!\! \max_{(G, \theta) \in \mathcal{P}^1} C(\Prob^G_\theta \given \cD)\,,
\end{equation}
where $\mathcal{P}^1\defi \mathcal{G}^1\times \Theta$ and $\mathcal{G}^1 \subsetneq \mathcal{G}$ is the set of DAGs which contain no edge of the form\footnote{To the best of our knowledge, we are the first who extend/adapt the setting suggested in \citep{lerner2001exact} to do probabilistic multi-dimensional classification when targeting the (regularized) joint conditional likelihood function.} $Y \longrightarrow X$.
\end{proposition}

We assume in this document that parameter learning can be optimally solved. In general, this is a strong assumption. However, we often deal with factorizations of $\Prob$ where each factor involves a small number of variables. In these cases, we hope one can learn the parameters well (certainly much better than in a global model). This is a condition we expect from local models in the factorization in order to prove the optimality of the framework. Note that the cardinality $|\mathcal{G}^1| = R(K) 2^{KQ} R(Q)$ can be much smaller than $|\mathcal{G}| = R(K+Q)$, where $R(\cdot)$ is Robinson's formula \citep{bielza2011multi}. Thus, looking for the best $(G, \theta)$ over $\mathcal{P}^1$ can be much more practical than doing so over $\mathcal{P}$. The next proposition shows that finding an optimal pair $(G,\theta) \in \mathcal{P}^1$ is equivalent to finding an optimal pair whose $G$ contains no edge between features.
\begin{proposition} \label{pro:Decomposability}
For any $G \in \mathcal{G}^1$, the joint conditional distribution \eqref{eq:jointConditionalProbabilities} can be factorized (according to $G$):
\begin{equation}\label{eq:CJPD_BND}
    \Prob^G_\theta(\by \given \bx) = \prod_{Y \in \mathbf{Y}} \Prob_\theta \left(y \given \pi_y\right) \,, \forall (\bx,\by) \in \cX \times \cY \,,
\end{equation}
where $\pi_y$ is the configuration for the parents of $Y$ (according to $G$) that is compatible with $(\bx,\by)$. Moreover, the following relation holds:
\begin{equation}\label{eq:Learning_problem_BND_simplified}
   \max_{(G,\theta) \in \mathcal{P}^1} C(\Prob^G_\theta\given \cD) = \max_{(G,\theta) \in \mathcal{P}^2} C(\Prob^G_\theta\given \cD) \,,
\end{equation}  
\noindent
where $\mathcal{P}^2 \defi \mathcal{G}^2\times \Theta$ and $\mathcal{G}^2 \subsetneq \mathcal{G}^1$ consists of $R(K) 2^{KQ}$ DAGs with no edges between any two elements of $\bX$.
\end{proposition}

Thus, we formulate the new optimization problem:
\begin{align}\label{eq:Learning_problem_BND_Final}
\Prob^{G^*}_{\theta^*}: ~ (G^*,\theta^*) &\in \argmax_{(G,\theta) \in \mathcal{P}^2} \log \prod_{n=1}^N \prod_{Y \in \mathbf{Y}} \Prob_{\theta}\left(y_n \given \pi_{y_n}\right)\,.
\end{align}
It is clear that solving \eqref{eq:Learning_problem_BND_Final} may lead to sub-optimal solutions, compared to solving \eqref{eq:Learning_problem_BN} if the assumption that the parameter learning problem is optimally solved does not hold, and in that case the relation $\mathcal{G}^2\subsetneq \mathcal{G}$ implies that the best CLL score attained over $\mathcal{G}^2$ is at best the one attained over $\mathcal{G}$. However, there are strong motivations for why one should solve \eqref{eq:Learning_problem_BND_Final} in practice, instead of \eqref{eq:Learning_problem_BN}.

First, the optimality of \eqref{eq:Learning_problem_BND_Final} can be reachable under milder conditions, while the optimality of \eqref{eq:Learning_problem_BN} is often unreachable. In fact, solving \eqref{eq:Learning_problem_BN} is often impractical because optimizing the CLL function can be impractical even if $G \in \mathcal{G}$ is given \citep{friedman1997bayesian}. However, one can be much more optimistic about solving \eqref{eq:Learning_problem_BND_Final}. As will be shown in Section \ref{sec:Algorithmic_Solution}, solving \eqref{eq:Learning_problem_BND_Final} is possible as long as one can learn a set of (independent) probabilistic classifiers, plus learning an optimal DAG over the class variables. So one can use all current/future developments of both probabilistic classification and graphical model learning towards solving \eqref{eq:Learning_problem_BND_Final}. 

Second, as will be shown in Section \ref{sec:Algorithmic_Solution}, $\forall G \in \mathcal{G}^2$ and $\forall \bx \in \cX$, $\Prob^G_{\theta}(\mathbf{Y} \given \bx)$ can be factorized as a product of conditional probability distributions whose conditional part is always specified by a multivariate continuous variable. This provides us with a rich representational capacity as discussed in Section \ref{sec:Representational_Capacity}. In particular, any probabilistic classifier can be directly employed to model conditional probability distributions without requiring any data preprocessing transformation, leading to a rich framework for the employment of sophisticated techniques. The representational capacity would be much weaker if one had to parameterize $G \in \mathcal{G}\setminus \mathcal{G}^1 \supsetneq \mathcal{G}\setminus \mathcal{G}^2$ because it would be needed to find some parametric model to encode all conditional density functions $\Dens^G_\theta(z\given \pi_z)$ whose conditional part would be specified by a mixture of discrete and continuous variables. This would be a challenging problem by itself, especially if one does not want to use any data preprocessing transformation either before or during the training phase.

Our final simplification of the optimization problem while keeping optimality is to realize that we can seek for an optimal $G$ where all continuous variables are parents of every class variable, that is, $\mathbf{X}_c \subset \Delta_G^Y,~ \forall Y\in \mathbf{Y}$. 
Besides being non-restrictive (we are forcing arcs to stay put, hence we can always fit any ``simpler'' distribution which would have dropped some connections by the appropriate parameter learning), this condition has also a positive consequence, as it allows us to use methods which are not able to handle mixed setups of continuous and discrete variables. 

Therefore, we introduce an updated version of \eqref{eq:Learning_problem_BND_Final} in which we only force the global learning algorithm to explicitly handle the discrete features, while assuming all continuous ones are passed on to the learning of local models. More formally, let $\mathcal{G}^3 \subsetneq \mathcal{G}^2$ be the set of $R(K)2^{K|\mathbf{X}_d|}$ DAGs such that, $\forall G\in \mathcal{G}^3$ and $\forall Y\in \mathbf{Y}$, we have $\mathbf{X}_c \subset \Delta_G^Y$. We formulate the optimization problem as:
\begin{equation}\label{eq:Learning_problem_BND_relaxed}
   \Prob^{G^*}_{\theta^*}: ~ (G^*,\theta^*) \in \argmax_{(G,\theta)\in \mathcal{P}^3}  \log \prod_{n=1}^N \prod_{Y \in \mathbf{Y}} \Prob_{\theta}\left(y_n \given \pi_{y_n}\right)\,,
\end{equation}
where $\mathcal{P}^3 = \mathcal{G}^3 \times \Theta$. 
\begin{proposition}\label{pro:Learning_problem_P3}
Assume the parameter learning problem is optimally solved. 
The following relation holds 
\begin{equation}\label{eq:Learning_problem_BND_simplified_vs_relaxed}
   \max_{(G,\theta) \in \mathcal{P}^2} C(\Prob^G_{\theta} \given \cD) = \max_{(G,\theta) \in \mathcal{P}^3} C(\Prob^G_{\theta} \given \cD)\,.
\end{equation} 
\end{proposition}
The conclusion here is that we can have a globally optimal probabilistic MDC whose optimization is done via~\eqref{eq:Learning_problem_BND_relaxed}, potentially saving significant time and data requirements for training. One needs ``only'' to learn the local conditional models (factors) of the expression, so long as we have an efficient solver to find the DAG $G$ inducing a good factorization. Moreover, we hope for a valid (in terms of being an I-map for the true distribution~\citep{bouckaert1994properties,koller2009probabilistic}) yet simple $G$. Hence, in the next section, we show that solving \eqref{eq:Learning_problem_BND_relaxed} can be optimally decomposed into learning a set of probabilistic classifiers and learning an optimal DAG. 

\subsection{Algorithmic Solution}\label{sec:Algorithmic_Solution}

In order to solve \eqref{eq:Learning_problem_BND_relaxed}, we first need to model the local conditional probability distributions:
\begin{equation}\label{eq:Local_Distributions}
     \Prob_{\theta}\left(Y \given \Delta_G^Y \right) \,, \forall G \in \mathcal{G}^3 \,, \forall Y \in \mathbf{Y}\,.  
\end{equation}
Given $G$, for any $Y \in \mathbf{Y}$, let $\Delta^Y_d = \Delta^Y_G \setminus \mathbf{X}_c$ be the set of all discrete variables in $\Delta^Y_G$. Let $\Pi^Y_d$ be the set of all configurations of $\Delta^Y_d$. Hence, each local distribution \eqref{eq:Local_Distributions} is represented by $|\Pi^Y_d|$ distributions \begin{equation}\label{eq:Local_Classifiers}
     \Prob_{\theta}\left(Y \given \pi, \mathbf{X}_c \right) \,, \forall \pi \in \Pi^Y_d\,.  
\end{equation} 

Thus, the optimization problem \eqref{eq:Learning_problem_BND_relaxed} becomes 
\begin{align}\label{eq:Learning_problem_BND_relaxed_reformulated}
   (G^*,\theta^*) &\in
  \argmax_{(G,\theta)\in \mathcal{P}^3} \sum_{Y \in \mathbf{Y}} 
 \sum_{\pi \in \Pi^Y_d} \log \prod_{(\bx,\by) \in \cD_\pi}  \Prob_{\theta}\left(y \given \pi, \bx^c\right),\nonumber
\end{align}
with $\cD_\pi \defi \{(\bx,\by) \in \cD | \pi^d_y = \pi \}$. A key point is the separation of discrete conditionals $\pi$ and continuous conditionals $\bx^c$. Such separations were used in learning BNs optimizing the likelihood function \citep{atienza2022hybrid}
Moreover, we have 
\begin{equation} \label{eq:reformulated_problem}
\max_{(G,\theta)\in \mathcal{P}^3} C(\Prob^G_\theta \given \cD) =\max_{G\in \mathcal{G}^3} \sum_{Y \in \mathbf{Y}} 
 \sum_{\pi \in \Pi^Y_d} \max_{\theta \in \Theta} C(\Prob_\theta\given Y,\pi,\cD)\,,
\end{equation}
where $$C(\Prob_\theta\given Y,\pi,\cD)=\log \prod_{(\bx,\by) \in \cD_\pi}  \Prob_{\theta}\left(y \given \pi, \bx^c\right).$$
This means that we can reformulate the optimization problem \eqref{eq:Learning_problem_BND_relaxed} as a two-phase optimization problem: (P1) for any tuple $(Y,\pi) \in \mathbf{Y} \times \Pi^Y_d$ (considering the possible $\Delta^Y_d$), learn the optimal parameter set $\theta^*$ of each distribution \eqref{eq:Local_Classifiers} which optimizes the local CLL function, i.e., 
\begin{align}\label{eq:P1}
    \theta^*_{Y,\pi} &\in \argmax_{\theta \in \Theta} C(\Prob_\theta\given Y,\pi,\cD),
\end{align}
and then (P2) learn the best DAG $G^* \in \mathcal{G}^3$ which maximizes the CLL function: 
\begin{equation}
G^* = \argmax_G C(\Prob^G_{\theta^*} \given \cD )\,,
\end{equation}
where 
\begin{equation}\label{eq:P2}
    C(\Prob^G_{\theta^*} \given \cD ) =  \sum_{Y \in \mathbf{Y}} 
 \sum_{\pi \in \Pi^Y_d} C(\Prob_{\theta^*_{Y,\pi}}\given Y,\pi,\cD).
\end{equation}
Problem (P1) can be solved for each  tuple $(Y,\pi) \in \mathbf{Y} \times \Pi^Y_d$, for each possible $\Delta^Y_p\in\mathcal{F}^Y$ independently (where $\mathcal{F}^Y$ is a set of candidate parent sets for $Y$). (P2) can be cast as the structure learning for BNs, so we can leverage the research on that topic \citep{kitson2023survey}. The elephant in the room here is the size of $\mathcal{F}^Y$ (for each $Y$), which will be discussed in Section~\ref{sec:Complexity_Learning_Problem}. 

In this paper, we solve \eqref{eq:P2} using GOBNILP \citep{bartlett2017integer,cussens2017bayesian} which is a state-of-the-art anytime globally optimal algorithm and can be easily adapted to handle regularized variants of CLL function as presented in Section \ref{sec:Complexity_Learning_Problem}. Intuitively, GOBNILP, which was designed for generative learning of Bayesian networks, can be instead used to reformulate the problem (P2) as learning a collection of parent sets $\{\Delta^Y_d: Y\in\bY\}$ which optimizes the CLL function \eqref{eq:P2} and together satisfy the DAG properties. It uses the local scores: $\forall Y\in\bY, \forall\Delta^Y_p\in\mathcal{F}^Y$:
\begin{align} \label{eq:local_score_for_Y}
C(Y, \Delta^Y_d) = \sum_{\pi \in \Pi^Y_d} C(\Prob_{\theta^*_{Y,\pi}}\given Y,\pi,\cD) \,,
\end{align}
where we simplified the notation by removing $\theta$ and $\cD$, since parameters have been already learned via~\eqref{eq:P1} and data are fixed. Problem (P2) can be expressed as an Integer Programming (IP) problem:
\begin{align}
    \maximize  & \sum_{Y\in \mathbf{Y}}  \sum_{\Delta^Y_d\in \mathcal{F}^Y}  \gamma(\Delta^Y_d) \cdot C(Y, \Delta^Y_d)    \,, \label{eq:target_function_GOBNILP}\\
     \text{Subject} \text{ to } &
     \sum_{\Delta^Y_d\in \mathcal{F}^Y} \gamma(\Delta^Y_d) =1 \,, \forall Y \in \mathbf{Y}  \, ,\nonumber \\
     &\sum_{Y \in \mathbf{Y}'} \!\sum_{\substack{\Delta^Y_d \in \mathcal{F}^Y \\ \Delta^Y_d \cap \mathbf{Y}' = \emptyset}} \! \!\!\! \! \gamma(\Delta^Y_d) >1  \,, \forall \mathbf{Y}' \subseteq \mathbf{Y}\,, |\mathbf{Y}'| >1 \, , \nonumber\\
     &\gamma(\Delta^Y_d)  \in \{0,1\} \,, \forall Y \in \mathbf{Y}, \forall ,\Delta^Y_d \in \mathcal{F}^Y \,.\nonumber
\end{align}  

The implementation is given in Algorithm \ref{alg:learn_GBNC}, which returns a $(G^*, \theta^*) \in \mathcal{P}^3$ of \eqref{eq:Learning_problem_BND_relaxed}. 
We call this type of model defined by $(G^*, \theta^*)$ a generalized Bayesian Network classifier (\GBNC{}).
Note that the loops starting in lines $2$ and $3$ can be easily parallelized since the local distributions \eqref{eq:Local_Classifiers} can be learned independently.

	\begin{algorithm} [!ht]
	\caption{Learning a \GBNC{} of \eqref{eq:Learning_problem_BND_relaxed}}\label{alg:learn_GBNC}
	\begin{algorithmic}[1]
   \STATE {\bfseries Input:} Data $\mathcal{D}$, Probabilistic hypothesis spaces. \;
   \FOR{$Y \in \mathbf{Y}$}
        \FOR{$\Delta^Y_d \in \mathcal{F}^Y$}
            \FOR{$\pi \in \Pi^Y_d$}
                \STATE Solve \eqref{eq:P1} and store it in a proper data structure
            \ENDFOR
            \STATE Compute $C(Y,\Delta^Y_d)$ by \eqref{eq:local_score_for_Y} using stored values\;
        \ENDFOR
   \ENDFOR
   \STATE Find a best collection $\{\Delta^Y_d:~ Y \in \mathbf{Y}\}$ which optimizes \eqref{eq:target_function_GOBNILP} using GOBNILP \;
   \STATE {\bfseries Output:} A \GBNC{} $(G^*, \theta^*) \in \mathcal{P}^3$ of \eqref{eq:Learning_problem_BND_relaxed} \;
   \end{algorithmic}
   \end{algorithm}

The optimality of the proposed framework can be derived as a consequence of Proposition \ref{pro:Learning_problem_BND}--\ref{pro:Learning_problem_P3}.
\begin{corollary}\label{cor:optimality}
Assume the chain rule of probability holds. Assume the parameter learning problem is optimally solved. The procedure to learn a classifier $(G^*, \theta^*)$ by Algorithm \ref{alg:learn_GBNC} is universal (for distributions in $\mathcal{P}^0$).
\end{corollary}

\subsection{Representational Capacity} \label{sec:Representational_Capacity}

To represent the joint conditional probability distribution $\Prob(\mathbf{Y} \given \mathbf{X})$, we need a set of probabilistic classifiers $\Prob': \cX_c \fromto \cY^k$ to estimate the local conditional probability distributions \eqref{eq:Local_Classifiers}. Local models $\Prob'$ are trained with what we call base learners. Note that discrete variables are not included in the input for $\Prob'$ (they are dealt with through the DAG optimization), which also facilitates learning and representational capacity.

First, it allows us to represent the distribution $\Prob(\mathbf{Y} \given \mathbf{X})$ where $\mathbf{X}$ can contain both continuous features and discrete features without requiring any preprocessing transformation either before or during the training phase. We never face the problem of representing qualitative data for use as input as deep learning does \citep{hancock2020survey}. Besides, representing qualitative data for use as input is arguably the most critical obstacle for generalizing Classifier Chains (CCs) \citep{dembczynski2010bayes,read2021classifier}, which is a state-of-the-art multi-label classification framework, to cope with MDC.  Moreover, we naturally overcome a bottleneck in the development of Multi-dimensional Bayesian network classifiers (MDBNCs) \citep{gil2021multi} that is a shortage of classifiers for the cases of continuous features, and mixed features. 

Second, the probabilistic classifier inducing $\Prob'$ can be freely chosen according to our needs. It can be as intuitive as $k$-NN classifiers \citep{cover1967nearest} and can be as counter-intuitive as ensembles of deep networks \citep{ganaie2022ensemble}. This allows us to employ sophisticated probabilistic classifiers to encode complex probabilistic relationships within $\Prob'_{Y,\pi}\defi\Prob_{\theta}\left(Y \given \pi, \mathbf{X}_c \right)$, $\forall \pi \in \Pi^Y_d$. For example, when each image is encoded using an $\bx$, a convolutional network \citep{lecun2015deep} can be employed to encode $\Prob'_{Y,\pi}$. If one seeks for more accurate \GBNCs, there should be no restriction on the use of ensemble learning methods, except the availability of computational resources. This flexibility of the framework is remarkably different from existing probabilistic MDC approaches~\citep{gil2021multi,jia2022multi}. Roughly speaking, so long as you train good local models $\Prob'_{Y,\pi}: \cX_c \fromto \cY^k$ (for which you can use all toolsets available in the literature for ``standard'' single-class-variable classification), the framework in this paper does the rest to combine them optimally into an MDC solution.

\subsection{Interpretability}\label{sec:Interpretability}

\GBNCs{} are interpretable at both the population and individual levels. At the population level, the structure $G$ provides a compact representation of the qualitative probabilistic relationships among feature and class variables. This graph representation is easy to interpret to end users when compared to an exponential number of masses provided by CP \citep{jia2021decomposition} and the (infinitely) many joint conditional distributions associated with the set of marginal probability distributions provided by BR \citep{jia2021decomposition}.     
At the individual level, the structure $G$ and its parameters specified by $\theta$ under the particular value of an individual $\bx$ form a compact representation of the qualitative and quantitative probabilistic relationships within $\Prob(\mathcal{Y}\given \bx)$, which can be seen as a BN over the class variables.  

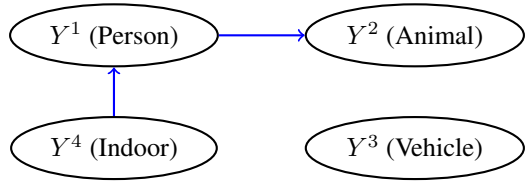
\begin{figure}[h!]
    \centering
    \begin{tikzpicture}[node distance=2.5cm, thick, main/.style = {draw, ellipse}]
        \node[main] (y1) at (0,0) {$Y^1$ (Person)}; 
        \node[main] (y2) at (4,0) {$Y^2$ (Animal)}; 
        \node[main] (y4) at (0,-1.5) {$Y^4$ (Indoor)}; 
        \node[main] (y3) at (4,-1.5) {$Y^3$ (Vehicle)}; 
        \draw[draw=blue,->] (y4) -- (y1);
        \draw[draw=blue,->] (y1) -- (y2); 
    \end{tikzpicture}  
\caption{A DAG over class variables learned from the PASCAL VOC 2007 data set.} \label{fig:DAG_PASCAL_VOC}
\end{figure}

As an example, we provide in Figure \ref{fig:DAG_PASCAL_VOC} a DAG over class variables learned from the PASCAL VOC 2007 data set whose description is given in Section \ref{sec:Experiments}.

\subsection{Regularization}
\label{sec:Complexity_Learning_Problem}

While Algorithm \ref{alg:learn_GBNC} helps to find an optimal \GBNC{} which maximizes the CLL function, the next proposition suggests that this best \GBNC{} may not always be the one we want, especially with regard to overfitting.  

\begin{proposition}  \label{pro:monotonicity}
Assume local models have parameters optimally learned.
Then $\forall Y\in \mathbf{Y}$ and $\forall \Delta, \Delta' \in \mathcal{F}^Y$ such that $ \Delta_d \subset \Delta'_d$, we have
\begin{align}\label{eq:monotonicity}
C(Y,\Delta_d) \leq C(Y,\Delta'_d) \,.
\end{align}
Therefore, at least one optimal solution of the Algorithm \ref{alg:learn_GBNC} is a fully connected DAG $G$. 
\end{proposition}  

Over-complex DAGs can happen frequently, especially when the local classifiers are learned without enforcing regularization terms. To seek for a better generalization, we propose a regularized variant of the CLL function:
\begin{align}\label{eq:regularized_CLL}
S(\Prob^G_{\theta} \given \cD ) = C(\Prob^G_{\theta} \given \cD ) - \sum_{Y \in \mathbf{Y}} \text{pen}(|\Delta^Y_d|, |\mathcal{D}|) \,,
\end{align}
where $\text{pen}(\Delta^Y_d, |\mathcal{D}|)$ can be the penalty term of any decomposable scoring function \citep{liu2012empirical}. Even a mild penalty can already help to reduce model complexity, but we leave this study to future work.

Algorithm \ref{alg:learn_GBNC} can be revised to learn \GBNCs{} of regularized variants \eqref{eq:regularized_CLL} as presented in Appendix \ref{sec:Naive_Algorithm_appendix} and \ref{sec:Refine_Algorithm_appendix}. 
Moreover, as shown in Appendix \ref{sec:Refine_Algorithm_appendix},
pruning rules \citep{de2018entropy} can be employed to find \GBNCs{} which optimize regularized variants \eqref{eq:regularized_CLL} without losing any optimality. This helps to greatly reduce the learning time because for each $Y \in \mathbf{Y}$, large candidate parent sets $\Delta^Y_d \in \mathcal{F}^Y$ are often pruned due to high penalties \citep{de2018entropy}. Finally, for a very large number of class variables, it is not unreasonable to expect the treewidth of the true distribution to be limited, so that one can bound the size of $\mathcal{F}^Y$ and use the scalability of (approximate) bounded-treewidth learning~\citep{scanagatta2016}. 

\section{Inference}
\label{sec:Inference_Problem}

The learned function $\Prob$ (defined via $G$ and $\theta$) provides, given an $\bx \in \cX$, a conditional joint probability distribution $\Prob(\cY \vert \bx)$ which is used to find the Bayes-optimal prediction (BOP) $\hat{\by}$ w.r.t a target loss function $\ell: \cY \times \cY \fromto \mathbb{R}_+$:
\begin{equation}\label{eq:BOP}
\hat{\by} \defi o(\Prob(\cY \vert \bx)) \in  \operatorname*{argmin}_{\overline{\by} \in \cY} \sum_{\by \in \cY} \ell(\by, \overline{\by}) \Prob(\by \given \bx)  \,  .
\end{equation}
Yet, different loss functions may call for different BOPs \eqref{eq:BOP} \citep{dembczynski2012label,gil2021multi,nguyen2021multilabel,waegeman2014bayes}. Knowledge about the probability distribution $\Prob(\cY \vert \bx)$ is necessary for finding BOP \eqref{eq:BOP} of any loss function. The complexity of finding BOP can greatly depend on the nature of the chosen loss function. This problem has been studied rarely in the MDC setting. An exception is \citep{bielza2011multi,gil2021multi}. Notably, in these works, finding BOP \eqref{eq:BOP} of some commonly used loss functions is shown to be equivalent to computing the most probable explanations (MPEs) of class variables when the classifier is an MDBNC. This is an interesting finding because it implies that the complexity of finding BOP \eqref{eq:BOP} depends on the nature of both the chosen loss function and the classifier. While this finding allows us to directly employ any current/future developments on exact/approximate MPE inference \citep{gil2021multi} to find BOP \eqref{eq:BOP} of some loss functions, one cannot get rid of the computational burden introduced by large numbers of features when working with MDBNCs. 

In our framework, we can also show that finding BOP \eqref{eq:BOP} of some loss functions is computing the MPEs of class variables. 
In the following, we describe the problem of finding BOP \eqref{eq:BOP} of two commonly used loss functions\footnote{We defer intensive studies on finding BOPs of other loss functions \citep{gil2021multi}[Section 4] to future work.} which are the \emph{Hamming loss} \eqref{eq:hamming} and the \emph{subset 0/1 loss} \eqref{eq:subset}: 
\begin{align}
\ell_H(\by, \hat{\by}) &\defi \frac{1}{K} \sum_{k=1}^K  \, \llbracket y^k \neq  \hat{y}^k \rrbracket \, , \label{eq:hamming} \\
\ell_S(\by, \hat{\by}) &\defi \llbracket \by \neq  \hat{\by} \rrbracket \, . \label{eq:subset}
\end{align}
The indicator $\llbracket A \rrbracket$ equals $1$ if the $A$ is true and $0$ otherwise. Thus, both losses generalize the standard $0/1$ loss in binary classification. As noted in \citep{bielza2011multi}, finding a BOP of $\ell_H$ and $\ell_S$ are respectively equivalent to finding $K$ marginals \eqref{eq:BOP_Hamming} and equivalent to finding one MPE \eqref{eq:BOP_Subset}:
\begin{align}
    \hat{y}^k & \in \operatorname*{argmax}_{\overline{y}^k \in \cY^k} \Prob(\overline{y}^k \given \bx) \,, \forall k \in [K] \,, \label{eq:BOP_Hamming} \\
    \hat{\by} &\in \operatorname*{argmax}_{\overline{\by} \in \cY} \Prob(\overline{\by} \given \bx) \,. \label{eq:BOP_Subset}
\end{align}

Hence, the model does not require retraining to allow for different BOP. Exact MPE and marginal inferences are NP-hard problems \citep{de2020almost,ROTH1996273,shimony1994finding}. However, in our framework, the complexity of MPE and marginal inferences only depend on the number of class variables. Thus, we do not encounter the computational burden introduced by large numbers of features, making the framework usable in practice in spite of that. Moreover, one can control the graph complexity among class variables by employing bounded-treewidth learning~\citep{NIE2017412}. 

\section{Experiments}\label{sec:Experiments}

This section presents a set of experiments to assess the usefulness of our proposal. 

\subsection{Experimental Setting}

We compare two instantiations of \GBNCs{} (\GBNC-S which optimizes \eqref{eq:regularized_CLL} and produces BOP \eqref{eq:BOP_Subset} of $\ell_S$, and \GBNC-H which optimizes \eqref{eq:regularized_CLL} and produces BOP \eqref{eq:BOP_Hamming} of $\ell_H$) with three probabilistic competitors found in the literature on $20$ tabular data sets \citep{jia2021decomposition} and one image data set \citep{everingham2010pascal}. The number of instances varies from $154$ to $28779$, the number of features varies from $10$ to $1536$, and the number of class variables varies from $2$ to $16$. It also contains $3$ data sets with mixed discrete and continuous features. 

We utilize an MDC version of the PASCAL VOC $2007$ data set \citep{everingham2010pascal}. We encode the objects found in that data set using $4$ class variables: Person (Yes and No), Animal (No animal, Bird, Cat, Cow, Dog, Horse and Sheep), Vehicle (No vehicle, Aeroplane, Bicycle, Boat, Bus, Car, Motorbike, Train) and Indoor (No indoor object, Bottle, Chair, Dining table, Potted plant, Sofa, TV/Monitor).

\begin{figure*}[ht!]
\centering
\begin{subfigure}[b]{0.98\linewidth}
\begin{subfigure}[b]{0.45\linewidth}
    \centering
    \includegraphics[width=\linewidth]{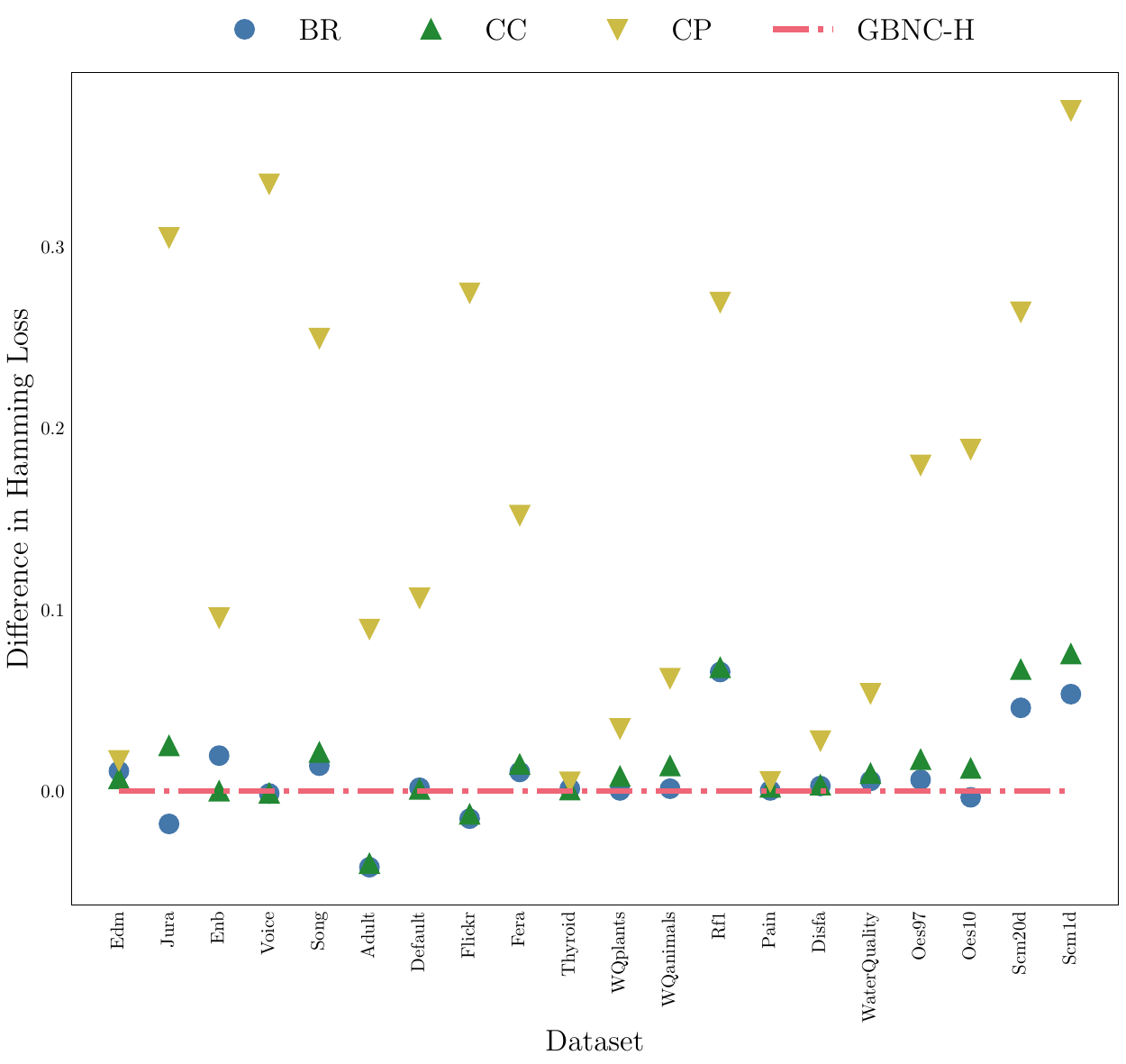}
\end{subfigure}
\hfill 
\begin{subfigure}[b]{0.45\linewidth}
    \centering
    \includegraphics[width=\linewidth]{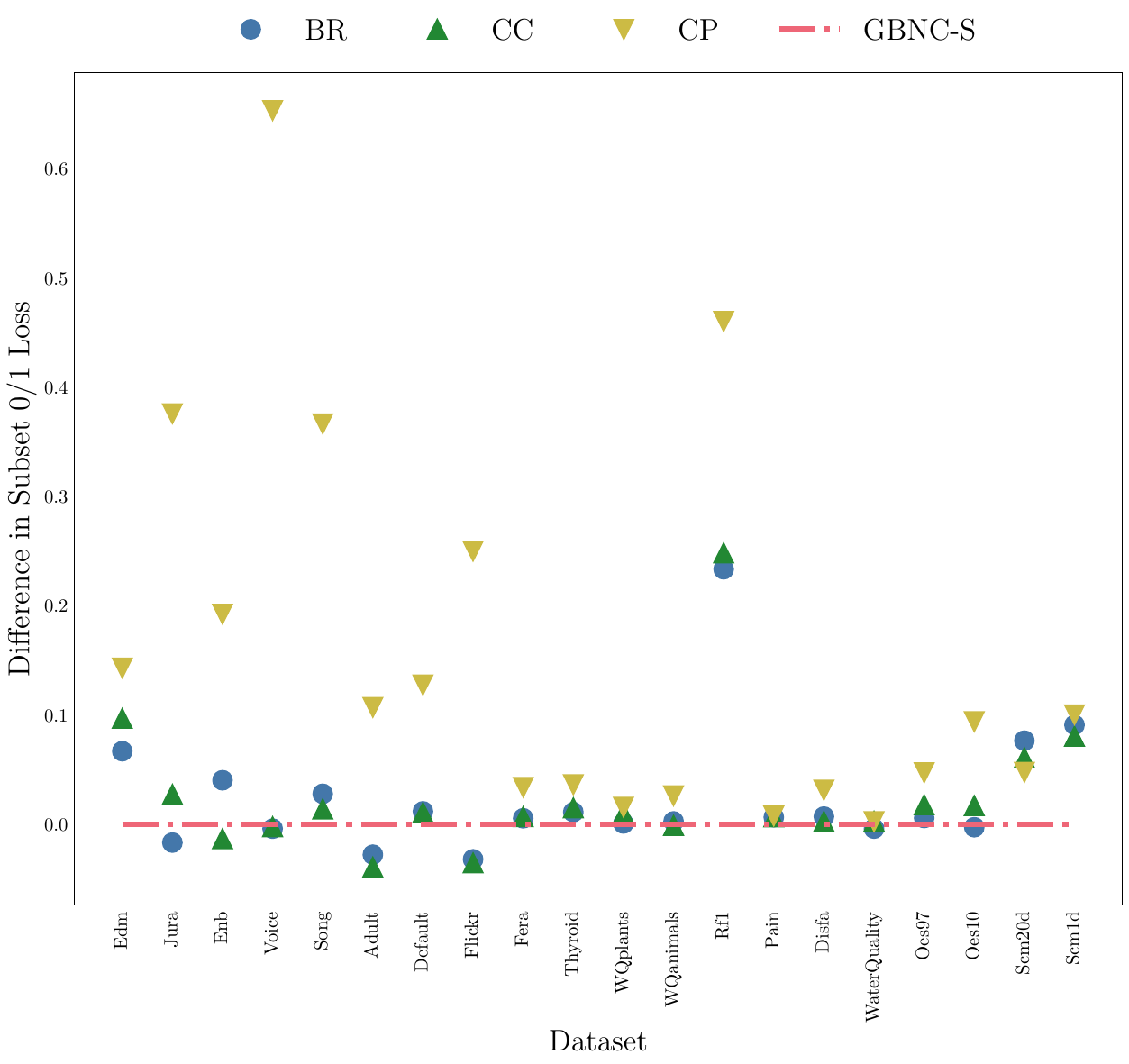}
\end{subfigure}
\caption{\small Base learner: Logistic Regression.}
\end{subfigure}

\hfill
\begin{subfigure}[b]{0.98\linewidth}
\begin{subfigure}[b]{0.45\linewidth}
    \centering
    \includegraphics[width=\linewidth]{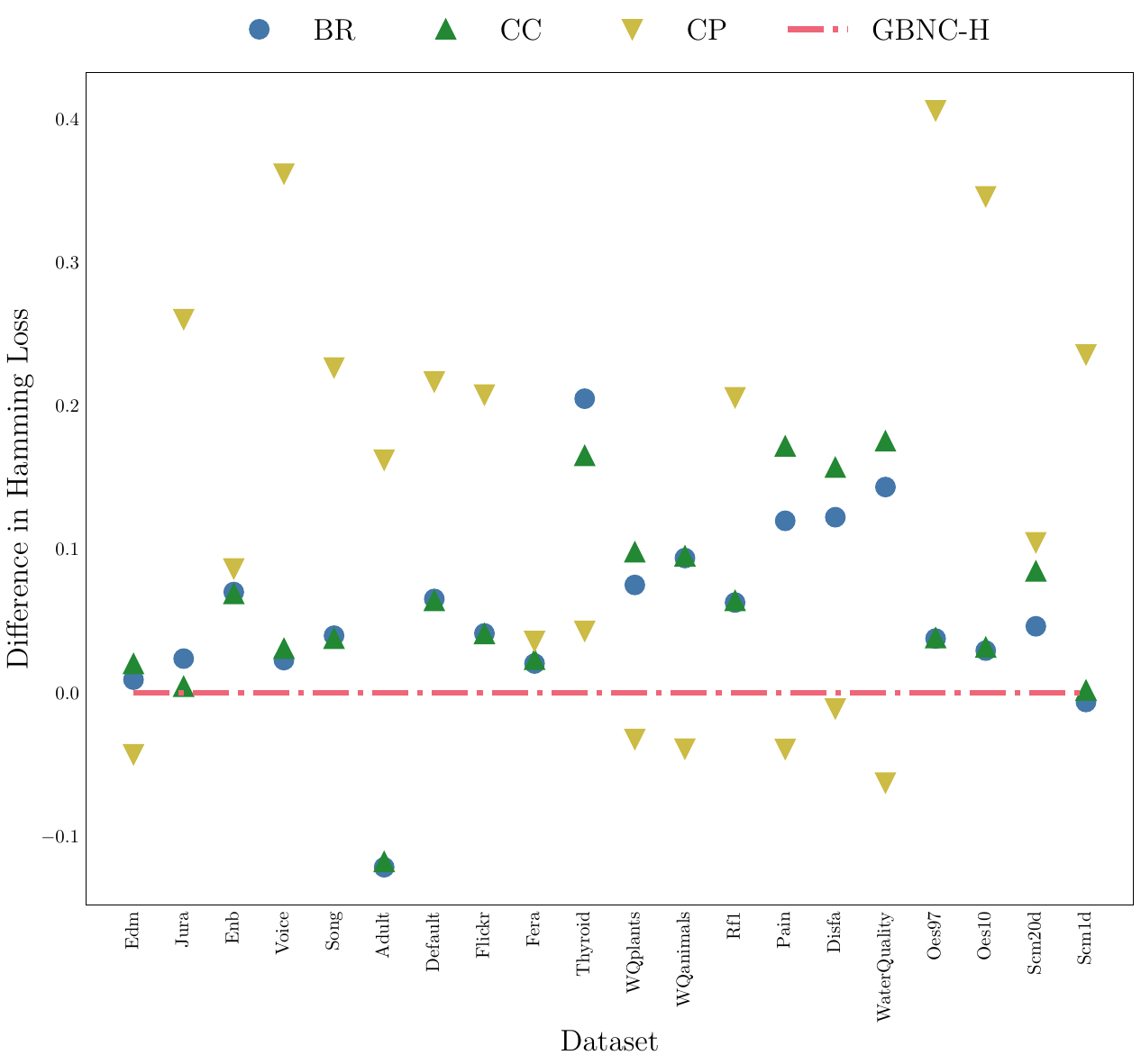}
\end{subfigure}
\hfill 
\begin{subfigure}[b]{0.45\linewidth}
    \centering
    \includegraphics[width=\linewidth]{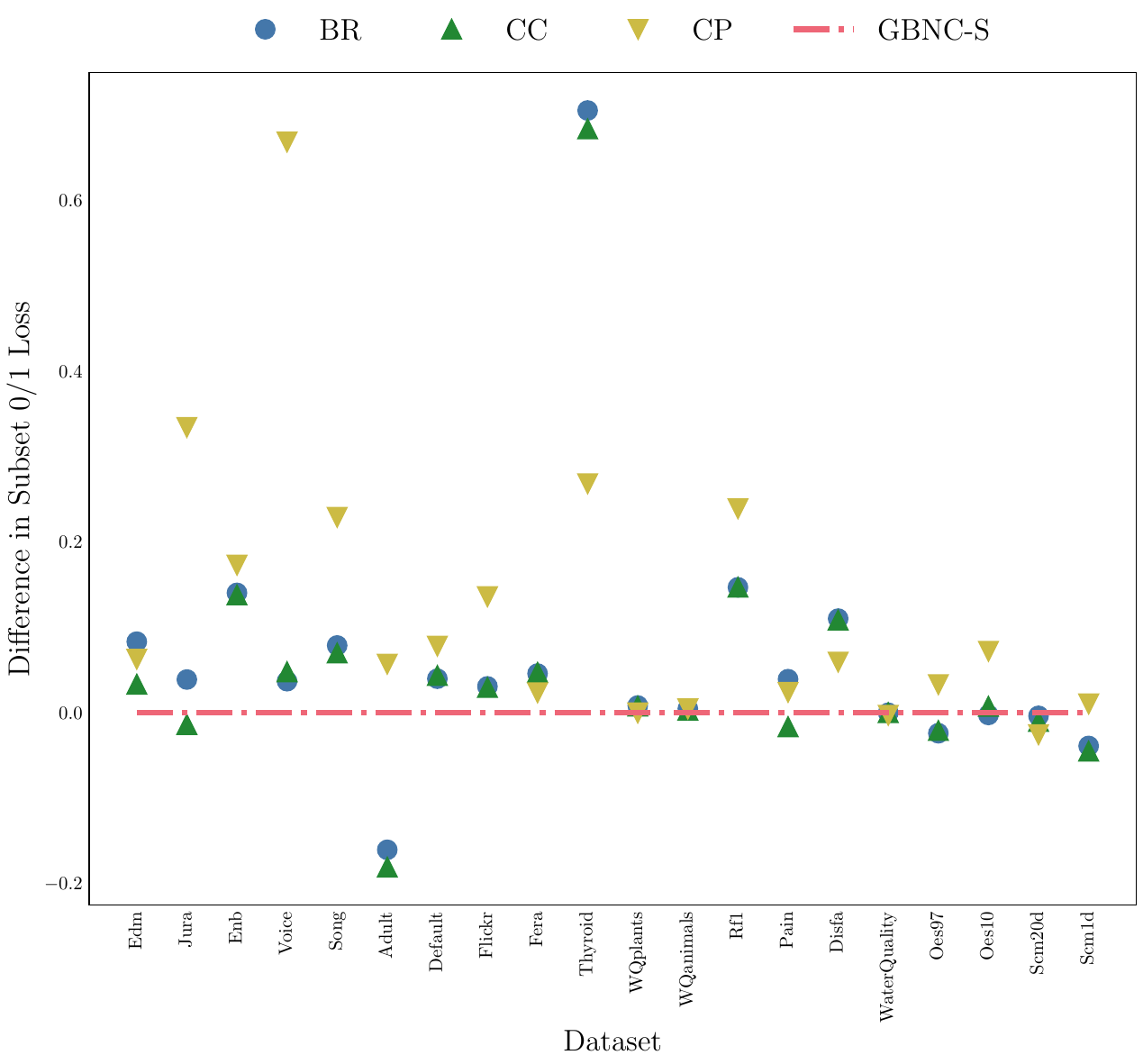}
\end{subfigure}
\caption{\small Base learner: Naive Bayes.}
\end{subfigure}
\caption{Tabular data sets: Performance differences to \GBNCs{} (negative means better than \GBNCs{}). Data sets (x-axis) are ordered by number of class variables.}
\label{fig:comparison}
\end{figure*}

For tabular data sets, we compare \GBNCs{} with BR and PC \citep{jia2021decomposition}[Section II], and CC \citep{jia2021decomposition}[Section III]. Because of the limitations of competitors to deal with mixed data, we follow the suggestion of \citep{jia2021decomposition} and convert discrete features/variables into continuous variables using one-hot encoding whenever they appear as parts of input of local classifiers of BR, PC and CC. Because we are not aware of any refinement of CC which can handle image data sets, we eliminate it from our comparison on the PASCAL VOC $2007$. For tabular data sets, we use logistic regression (LR) \citep{menard2002applied} and Naive Bayes (NB) classifiers \citep{domingos1996beyond} to estimate the local distributions \eqref{eq:Local_Classifiers} (one can use more complex models, but as we see in the remainder, these choices already yield state-of-the-art results, so we decided that further tuning would go beyond our scope). For the image data set, distributions \eqref{eq:Local_Classifiers} are estimated using ResNet-18 \citep{he2016deep} with the weights pre-trained on ImageNet \citep{deng2009imagenet}, which are calibrated using temperature scaling \citep{guo2017calibration}. Following the suggestion of \citep{zhang2017mixup}, we also employ \textit{mixup} to improve the generalization of ResNet-18. 

In our experiments, $\text{pen}(|\Delta^Y_d|, |\mathcal{D}|)$ is the penalty term of the Bayesian Information Criterion (BIC) \citep{schwarz1978estimating}. The experimental setting is detailed in Appendix \ref{sec:Experimental_Setting_appendix}. The source code has been made public at \url{https://github.com/yangyang-pro/probabilistic-mdc}.

\subsection{Results}

Overall, the results suggest the superiority of our framework against existing probabilistic MDC frameworks (See Table \ref{tab:res_imag_data}--\ref{tab:more_res_tab_data}, and Figure \ref{fig:comparison}). On the image data set, \GBNCs{} indeed provide the most promising $\ell_H$ and $\ell_S$ (See Table \ref{tab:res_imag_data}). 
\begin{table}[ht!]
  \centering
  \caption{Results (mean $	\pm	$ std.) on the image data set.}
\label{tab:res_imag_data}
\begin{tabular}{ccc}
\toprule 
\multicolumn{3}{c}{Hamming loss ($\ell_H$)}\\ \hline
\GBNC-H          &   BR &   CP\\ \hline
\bfseries 11.41 $	\pm	$ 0.35 &  12.51 $	\pm	$ 1.71  &    21.81 $	\pm	$ 7.62        \\   
 \hline
\multicolumn{3}{c}{Subset 0/1 loss ($\ell_S$)}\\ \hline
\GBNC-S          &   BR &   CP\\ \hline
 \bfseries 37.31 $	\pm	$ 0.84  & 41.57 $	\pm	$ 5.16   &  56.57 $	\pm	$ 13.28  \\
\bottomrule
\end{tabular}
\end{table}

\GBNCs{} yield the best average ranks over the $20$ tabular data sets, both for $\ell_H$ and $\ell_S$. Furthermore, Friedman tests \citep{demvsar2006statistical} on the ranks yield small p-values, and strongly suggest performance differences between the classifiers. We also conduct Nemenyi post-hoc test \citep{nemenyi1963distribution} and Conover post-hoc test \citep{conover1999practical,conover1979multiple} (see Table \ref{tab:more_res_tab_data}) to see if there are significant differences between pairs of classifiers. For each combination (among the $12$ combinations) of competitor, loss and local models, we find at least one test where \GBNCs{}  is significantly better than that competitor in almost all cases.

\begin{table}[ht!]
  \centering
  \caption{Average ranks and p-values of Friedman tests.} \label{tab:res_tab_data}
\begin{tabular}{l|ccccc}
\toprule 
\multicolumn{6}{c}{The cases of Hamming loss ($\ell_H$)}\\ \hline
Learner  &\GBNC-H &    BR & CC & CP &  p-value \\ \hline
LR  &  \bfseries 1.43  &  1.98  & 2.60   &  4.00      &     \bfseries   1.1e-09  \\
NB &   \bfseries 1.40 & 2.70   &  2.95  &    2.95    &     \bfseries   1.8e-04  \\ \hline 
\multicolumn{6}{c}{The cases of Subset 0/1 loss ($\ell_S$)}\\ \hline
Learner   & \GBNC-S   & BR & CC & CP & p-value \\ \hline
LR  & \bfseries 1.55   &  2.20  & 2.38   &   3.88     &    \bfseries 1.2e-07     \\
NB &   \bfseries 1.73 &   2.80 &  2.28  &   3.20     &     \bfseries 1.6e-03    \\
\bottomrule
\end{tabular}
\end{table}

\begin{table}[ht!]
  \centering
  \caption{Post-hoc tests: p-values.}
\label{tab:more_res_tab_data}
\begin{tabular}{l|cc|cc}
\toprule 
\multicolumn{5}{c}{The cases of $\ell_H$: p-values $< 0.05$ are given in bold}\\ \hline
\multirow{2}{*}{$H_0$ }&\multicolumn{2}{c|}{Nemenyi} &\multicolumn{2}{c}{Conover}\\ \cline{2-5}
         &   LR &   NB &   LR &   NB\\ \hline
\GBNC-H =  BR   & 0.529   & \bfseries 0.008    & 0.184 & \bfseries 0.002    \\
\GBNC-H =  CC   & \bfseries 0.021  & \bfseries 0.001  & \bfseries 0.006 &  \bfseries3.9e-04 \\  
\GBNC-H =  CP   & \bfseries 0.001   & \bfseries 0.001    & \bfseries4.6e-08  & \bfseries3.9e-04      \\ 
\hline 
BR =  CP       & \bfseries 0.001   & 0.9    & \bfseries 7.0e-07& 0.545      \\
CC =  CP       & \bfseries 0.003  & 0.9    & \bfseries 0.001 &  1  \\ 
BR =  CC       & 0.42   & 0.9    & 0.132 & 0.545    \\ 
\hline
\multicolumn{5}{c}{The cases of $\ell_S$: p-values $< 0.05$ are given in bold}\\ \hline 
\multirow{2}{*}{$H_0$ }&\multicolumn{2}{c|}{Nemenyi} &\multicolumn{2}{c}{Conover}\\ \cline{2-5}
                &   LR            &   NB &   LR &   NB\\ \hline
\GBNC-S =  BR   & 0.384          &   \bfseries 0.042   & 0.118     &  \bfseries 0.01   \\                
\GBNC-S =  CC   & 0.180          &  0.528     & \bfseries0.049    & 0.178   \\  
\GBNC-S =  CP   & \bfseries 0.001 & \bfseries 0.002     & \bfseries 4.9e-07     & \bfseries 5.6e-04   \\ 
\hline 
BR =  CP       & \bfseries 0.001 &  0.735     & \bfseries 1.4e-04     & 0.326 \\
CC =  CP       & \bfseries 0.001&   0.106    & \bfseries 5.5e-04     & \bfseries 0.026 \\ 
BR =  CC       & 0.9             &  0.563     & 0.671     & 0.198 \\ 
\bottomrule
\end{tabular}
\end{table}

Even if the Nemenyi post-hoc test may be too conservative, has low power, and may not detect existing differences when Friedman's test rejects the null hypothesis (as elaborated in \citep{ulacs2012cost} and also elsewhere), it already informs significant differences. Table~\ref{tab:more_res_tab_data} suggests that the use of both LR and NB as local models (i.e. base learners) leads to improvements with respect to other approaches. Actually, LR performs better with more class variables, while NB with fewer (these differences can be appreciated in the Appendices). Yet, it is not the goal of this work to answer this question. The experiments with two different local models (LR and NB) have the purpose of demonstrating the capabilities of the overall idea.

Our experimental results are in agreement with the results found in literature. First, CC can hardly be a state-of-the-art MDC approach \citep{jia2021decomposition}. Second, BR may provide competitive performance, especially when the number of class variables is not large \citep{wu2020multiNeurIPS}. On the other hand, our experiments suggest a very interesting result that \GBNC-H which estimates the joint conditional distribution and extracts marginal distributions using Definitions \eqref{eq:marginalConditionalProbabilities} often outperforms BR which directly estimates the marginal distributions. This suggests that capturing the dependency relationships can lead to more accurate estimates of the marginal probability distributions. 

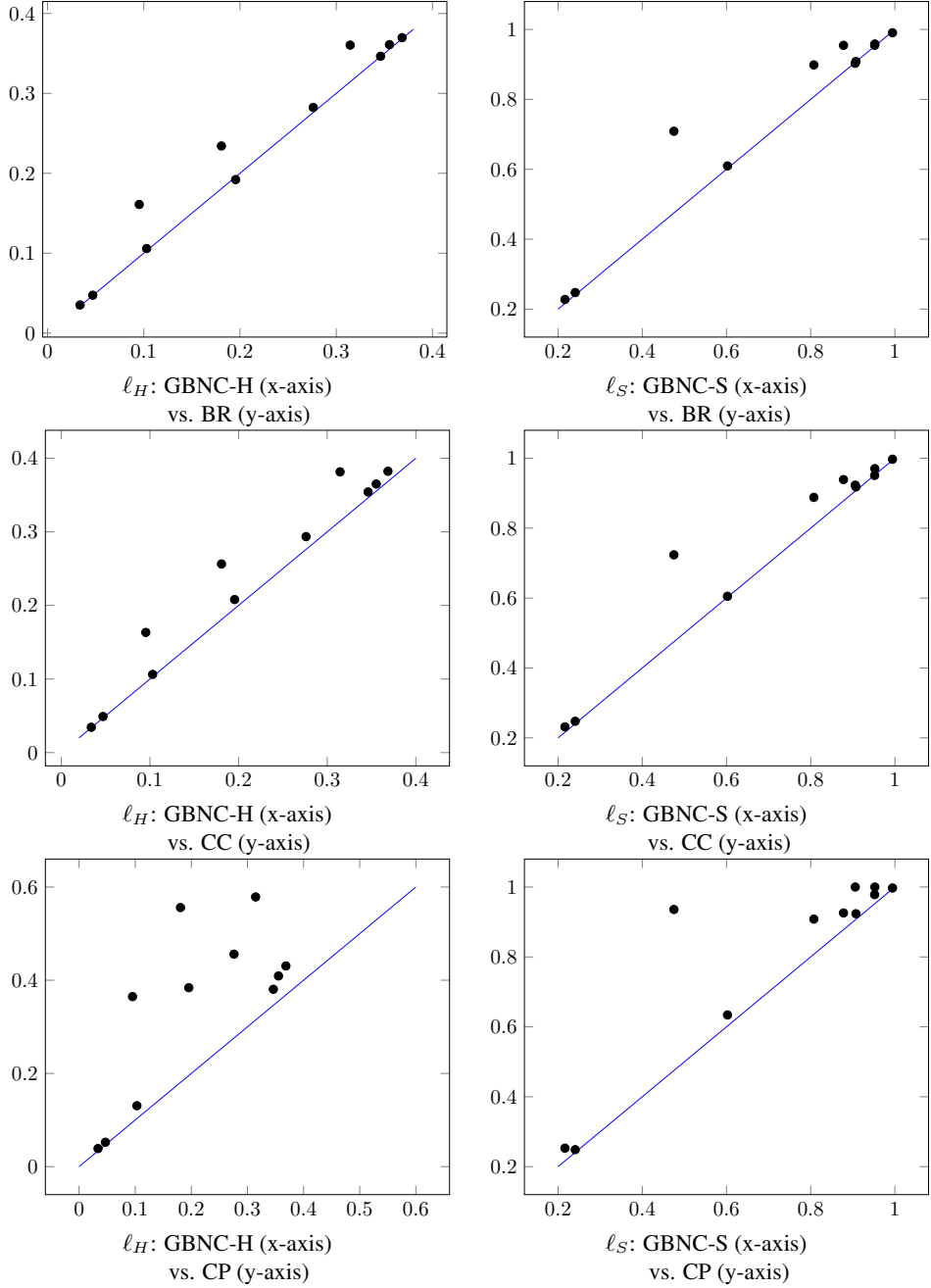
\begin{figure}[ht!]
\centering
\begin{tabular}{cc}
\begin{tikzpicture}[scale = 0.8]
\centering
\begin{axis}[%
scatter/classes={scatter src = explicit symbolic,%
    h={mark=*,draw=black},
    i={mark=*,draw=black},
    j={mark=*,draw=black},
    k={mark=*,draw=black},
    l={mark=*,draw=black},
    m={mark=*,draw=black},
    n={mark=*,draw=black},
    o={mark=*,draw=black},
    p={mark=*,draw=black},
    q={mark=*,draw=black},
    t={mark=*,draw=black}}]
\addplot[draw=blue,pattern=horizontal lines light blue]
 coordinates { (0.03,0.03) (0.38,0.38) };
\addplot[scatter,only marks,%
    scatter src=explicit symbolic]%
table[meta=label] {
x       y       label
0.3461	0.3465	h
0.3685	0.3698	i
0.0953	0.1610	j
0.0470	0.0474	k
0.1030	0.1058	l
0.3553	0.3610	m
0.2761	0.2824	n
0.1955	0.1921	o
0.3145	0.3604	p
0.1807	0.2342	q
0.0338	0.0352	t
    };
\end{axis}
\end{tikzpicture}
&
\begin{tikzpicture}[scale=0.8]
\centering
\begin{axis}[%
scatter/classes={scatter src = explicit symbolic,%
    h={mark=*,draw=black},
    i={mark=*,draw=black},
    j={mark=*,draw=black},
    k={mark=*,draw=black},
    l={mark=*,draw=black},
    m={mark=*,draw=black},
    n={mark=*,draw=black},
    o={mark=*,draw=black},
    p={mark=*,draw=black},
    q={mark=*,draw=black},
    t={mark=*,draw=black}}] 
\addplot[draw=blue,pattern=horizontal lines light blue]
 coordinates { (0.2,0.2) (1,1) };

\addplot[scatter,only marks,%
    scatter src=explicit symbolic]%
table[meta=label] {
x       y       label
0.9075	0.9085	h
0.9519	0.9547	i
0.4751	0.7088	j
0.2407	0.2474	k
0.6024	0.6096	l
0.9943	0.9906	m
0.9524	0.9584	n
0.9058	0.9033	o
0.8779	0.9546	p
0.8075	0.8986	q
0.2163	0.2276	t
    };
\end{axis}
\end{tikzpicture}
 \\
\small $\ell_H$: GBNC-H (x-axis)  & \small $\ell_S$: GBNC-S (x-axis)\\
\small  \quad   vs. BR (y-axis) &  \quad  \small  vs. BR (y-axis)
\\
\begin{tikzpicture}[scale=0.8]
\centering
\begin{axis}[%
scatter/classes={scatter src = explicit symbolic,%
    h={mark=*,draw=black},
    i={mark=*,draw=black},
    j={mark=*,draw=black},
    k={mark=*,draw=black},
    l={mark=*,draw=black},
    m={mark=*,draw=black},
    n={mark=*,draw=black},
    o={mark=*,draw=black},
    p={mark=*,draw=black},
    q={mark=*,draw=black},
    t={mark=*,draw=black}}]
\addplot[draw=blue,pattern=horizontal lines light blue]
 coordinates { (0.02,0.02) (0.4,0.4) };

\addplot[scatter,only marks,%
    scatter src=explicit symbolic]%
table[meta=label] {
x       y       label
0.3461	0.3542	h
0.3685	0.3823	i
0.0953	0.1633	j
0.0470	0.0491	k
0.1030	0.1063	l
0.3553	0.3650	m
0.2761	0.2935	n
0.1955	0.2080	o
0.3145	0.3815	p
0.1807	0.2563	q
0.0338	0.0344	t
    };
\end{axis}
\end{tikzpicture} 
&
\begin{tikzpicture}[scale=0.8]
\centering
\begin{axis}[%
scatter/classes={scatter src = explicit symbolic,%
    h={mark=*,draw=black},
    i={mark=*,draw=black},
    j={mark=*,draw=black},
    k={mark=*,draw=black},
    l={mark=*,draw=black},
    m={mark=*,draw=black},
    n={mark=*,draw=black},
    o={mark=*,draw=black},
    p={mark=*,draw=black},
    q={mark=*,draw=black},
    t={mark=*,draw=black}}]
\addplot[draw=blue,pattern=horizontal lines light blue]
 coordinates { (0.2,0.2) (1,1) };

\addplot[scatter,only marks,%
    scatter src=explicit symbolic]%
table[meta=label] {
x       y       label
0.9075	0.9179	h
0.9519	0.9509	i
0.4751	0.7236	j
0.2407	0.2476	k
0.6024	0.6051	l
0.9943	0.9972	m
0.9524	0.9703	n
0.9058	0.9230	o
0.8779	0.9389	p
0.8075	0.8881	q
0.2163	0.2315	t
    };
\end{axis}
\end{tikzpicture} 
\\
\small $\ell_H$: GBNC-H (x-axis)  & \small $\ell_S$: GBNC-S (x-axis)\\
\small  \quad   vs. CC (y-axis) &  \quad  \small  vs. CC (y-axis)
\\
\begin{tikzpicture}[scale=0.8]
\centering
\begin{axis}[%
scatter/classes={scatter src = explicit symbolic,%
    h={mark=*,draw=black},
    i={mark=*,draw=black},
    j={mark=*,draw=black},
    k={mark=*,draw=black},
    l={mark=*,draw=black},
    m={mark=*,draw=black},
    n={mark=*,draw=black},
    o={mark=*,draw=black},
    p={mark=*,draw=black},
    q={mark=*,draw=black},
    t={mark=*,draw=black}}]
\addplot[draw=blue,pattern=horizontal lines light blue]
 coordinates { (0,0) (0.6,0.6) };

\addplot[scatter,only marks,%
    scatter src=explicit symbolic]%
table[meta=label] {
x       y       label
0.3461	0.3806	h
0.3685	0.4307	i
0.0953	0.3649	j
0.0470	0.0524	k
0.1030	0.1308	l
0.3553	0.4092	m
0.2761	0.4559	n
0.1955	0.3840	n
0.3145	0.5787	p
0.1807	0.5559	q
0.0338	0.0389	t
    };
\end{axis}
\end{tikzpicture}
&
\begin{tikzpicture}[scale=0.8]
\centering
\begin{axis}[%
scatter/classes={scatter src = explicit symbolic,%
    h={mark=*,draw=black},
    i={mark=*,draw=black},
    j={mark=*,draw=black},
    k={mark=*,draw=black},
    l={mark=*,draw=black},
    m={mark=*,draw=black},
    n={mark=*,draw=black},
    o={mark=*,draw=black},
    p={mark=*,draw=black},
    q={mark=*,draw=black},
    t={mark=*,draw=black}}]
\addplot[draw=blue,pattern=horizontal lines light blue]
 coordinates { (0.2,0.2) (1,1) };

\addplot[scatter,only marks,%
    scatter src=explicit symbolic]%
table[meta=label] {
x       y       label
0.9075	0.9236	h
0.9519	0.9783	i
0.4751	0.9358	j
0.2407	0.2486	k
0.6024	0.6340	l
0.9943	0.9972	m
0.9524	1.0000	n
0.9058	1.0000	o
0.8779	0.9258	p
0.8075	0.9081	q
0.2163	0.2529	t
    };
\end{axis}
\end{tikzpicture}
\\
\small $\ell_H$: GBNC-H (x-axis)  & \small $\ell_S$: GBNC-S (x-axis)\\
\small  \quad   vs. CP (y-axis) &  \quad  \small  vs. CP (y-axis)
\\
 \end{tabular}
\caption{$\ell_H$ and $\ell_S$ with $K \geq 7$ (\textbf{base learner: \textit{LR}})}\label{fig:scatter_plots_K_more_than_7} 
\end{figure}

Although comparing ranks \citep{demvsar2006statistical} of classifiers is a common practice when one seeks short summaries of the performances, there is no golden rule about how the classifiers should be ranked. In this case, ranking the losses can not tell us whether there is any visible gain/loss. To gain more insights into the differences between classifiers, we make scatter plots for the losses provided by pairs of classifiers (See Figure \ref{fig:Hamming loss for pairs with LR}--\ref{fig:Subset 0/1 loss for pairs with NB} in Appendix \ref{sec:Results_appendix}). 
In all cases, \GBNC-H and \GBNC-S are rarely worse than others with visible differences, and visible gains of \GBNC-H and \GBNC-S are observed in all cases. Again, those figures suggest that \GBNC-H and \GBNC-S can consistently provide promising performance. 
In practice, we would expect to see approaches which take into account dependencies among the class variables brings more advantages when the number of class variables $K$ increases and the base learner is accurate. To show this ability of \GBNCs, we make scatter plots for the losses provided by pairs of classifiers on $11$ data sets with $K \geq 7$ with LR as the base learner (which is often more accurate than NB on these data sets). Figure \ref{fig:scatter_plots_K_more_than_7} confirms that \GBNCs{} indeed provide visible gains on these data sets. 

Finally, we acknowledge that one can devise creative ideas to tackle MDC indirectly via other approaches, so one might ask to which extent our experiments yield state-of-the-art performance in a broader sense. We emphasize that our goal is to improve on probabilistic MDC itself and to demonstrate the usefulness of this framework which has proven optimality properties and is very flexible to work with many other (off-the-shelf) classifiers as internal local models (i.e. base learners). If one embraces the framework and chooses strong local models, this is likely (based on the theoretical results) to perform very well for MDC.

\section{Conclusion}\label{sec:Conclusion}
We propose a formal framework for probabilistic multi-dimensional classification (MDC) in which learning an optimal multi-dimensional classifier can be decomposed into learning a set of probabilistic classifiers and learning an optimal Bayesian network (BN) structure. We discuss how single-class-variable probabilistic classification and BN learning can be directly integrated into the framework with respect to optimality, representational capacity and scalability. We present algorithmic solutions for the learning and inference problems and discuss on their complexity. Finally, a set of experiments highlights the usefulness of the MDC framework. We hope that this paper can open doors for further research on all these strongly related topics. 

\section*{Acknowledgements} 
This work was initiated when all authors were at the TU Eindhoven. Vu-Linh Nguyen has been funded by the Junior Professor Chair in Trustworthy AI (Ref. ANR-R311CHD). 
Yang Yang has been funded by the Research Foundation – Flanders (FWO, G097720N).
Cassio de Campos thanks the support of EU European Defence Fund Project KOIOS (EDF-2021-DIGIT-R-FL-KOIOS) and Dutch NWO Perspectief 2022 Project PersOn (P21-03).
\bibliographystyle{abbrvnat}
\bibliography{PMDC.bib}

\newpage

\appendix
\section{Notation and acronyms}
\begin{table}[H]
\begin{center}
\caption{Notation and acronyms}
\begin{tabular}{ll}
\hline
symbol/acronym & meaning \\
\hline
$\cX$, $\bx$ & instance space, instance \\
$\cY$, $\by$ & output space, outcome \\
$X^p$, $Y^k$ & feature, class variable\\
\hline
$K$, $Q$ & number of class variables, number of features\\
$\llbracket \cdot \rrbracket$ & indicator function \\
$[n]$ & set $\{ 1,\ldots, n\}$ of natural numbers \\
\hline
$\Prob(\by \given \bx)$ & probability of outcome $\by$ given $\bx$ \\
$\Prob(y^k \given \bx)$ & marginal probability of relevance for outcome $Y^k = y^k $ given $\bx$\\ 
\hline
$\cD$ & training data \\
\hline
$\ell$, $\ell_S$, $\ell_L$ & MDC loss function, Subset 0/1 loss, Hamming loss\\
\hline
$G$, $\theta$ & Structure (i.e., a DAG) of a BN, Parameter set of a BN \\
$\Delta^Y_G$ & Parent set of $Y$ in $G$ \\
$\Pi^Y_d$ & Set with all possible configurations of the discrete parents of $Y$\\
$\pi$ & Configuration of the parents of a variable, stored as pairs (variable, value)\\
$\mathcal{G}$ & Set of the $R(K+Q)$ possible DAGs \\
$\mathcal{G}^1$ & Set of the $R(K) 2^{KQ} R(Q)$ DAGs which contain no edge of the form $Y \longrightarrow X$\\  
$\mathcal{G}^2$ & Set of the $R(K) 2^{KQ}$ DAGs, whose elements contain no edge between features\\
$\mathcal{G}^3$ & Set of the $R(K)2^{K|\mathbf{X}_d|})$ DAGs such that, $\forall G\in \mathcal{G}^3$ and $\forall Y\in \mathbf{Y}$, we have $\mathbf{X}_c \subset \Delta^Y$\\
\hline
DAG & Directed acyclic graph\\
MDC & Multi-dimensional classification\\
BN & Bayesian network\\
BOP & Bayes-optimal prediction\\
CP & class powerset \\
CCs & classifier chains \\
BR & binary relevance \\
\hline
\end{tabular}
\end{center}
\end{table}

\section{Proofs of Propositions}\label{sec:Proofs_of_Propositions_appendix}

This section presents proofs for the propositions stated in the main paper. When it is necessary, we recall related notions and results in the literature before presenting proofs. 

\subsection{Proposition \ref{pro:Learning_problem_BND}}

We first present Lemmas which are necessary to complete the proof of Proposition \ref{pro:Learning_problem_BND}.

\begin{lemma}\label{lem:I-maps_2_CLL}
Assume there is a $G'\in \mathcal{G}$ such that $G'$ is an I-map for $\Prob \in \mathcal{P}^0$. All the I-maps $G$ of $\Prob$ induce the same CLL score \eqref{eq:CLL}.
\end{lemma}
\begin{proof}
Reminding that conditional joint probability distribution is 
\begin{align}
    \Prob(\by \given \bx) \defi \frac{\Prob(\bx, \by)}{\sum_{\by \in \cY} \Prob(\bx, \by)} \,, \forall (\bx,\by) \in \cX \times \cY \,. 
\end{align}
Assume $G \in \mathcal{G}$ is an I-map of $\Prob$. Because of that, we have
$\Prob(\bx,\by) =  \Prob^G_\theta(\bx, \by)$ for a certain $\theta$, with factorization respecting $G$. This implies
\begin{align}
    \Prob(\by \given \bx) = \frac{\Prob(\bx, \by)}{\sum_{\by' \in \cY} \Prob(\bx, \by')} = \frac{\Prob^G_{\theta}(\bx, \by)}{\sum_{\by' \in \cY} \Prob^G_{\theta}(\bx, \by')} =  \Prob^G_{\theta}(\by \given \bx) \,, \forall (\bx,\by) \in \cX \times \cY \,. 
\end{align}
Then it is clear that
\begin{align*}
    C(\Prob \given \cD) =  \log \prod_{n=1}^N \Prob(\by_n \given \bx_n)  =  \log \prod_{n=1}^N \Prob^G_{\theta}(\by_n \given \bx_n) = C(\Prob^G_{\theta} \given \cD)\,.
\end{align*}
Thus, all the I-maps $G$ of $\Dens$ have the same CLL score on $\cD$. 
\end{proof}

\begin{lemma}\label{lem:I-map_in_G'}
Assume elements of $\mathbf{X}$ are always made available. Assume there is a $G\in \mathcal{G}$ such that $G$ is an I-map for $\Prob \in\mathcal{P}^0$. Then, there is at least one $G'\in \mathcal{G}^1$ which is an I-map for $\Prob$. 
\end{lemma}
\begin{proof}

As long as the chain rule of probability is valid, we can lazily pick up any topological ordering $t'(1), \ldots, t'(K +Q)$ on $\mathbf{Z} = \mathbf{X} \cup \mathbf{Y}$ in which the $Q$ features occupy the first $Q$ places (and the $K$ class variables occupy the next $K$ places) and add arcs from each feature/variable to the ones succeed it until having a fully connected DAG $G'$. It is clear that $G'\in \mathcal{G}^1$ because we never add any arc of the form $Y\longrightarrow X$. Moreover, $G'$ is an I-map\footnote{While this is already enough to complete the proof, fully connected DAGs are not really the goal of learning BNs. Sparser I-maps can be easily constructed by only adding arcs which preserve conditional dependencies when following the ordering. Also, during the execution of the main algorithms that we use, we naturally find small graphs because of the penalizations that are used (similar guarantees as those that exist for learning BNs).} of $G$ (and $\Prob$) because $\mathcal{I}(G') = \emptyset$. 
\end{proof}

In the following, we present a proof of Proposition \ref{pro:Learning_problem_BND}. 
\begin{proof}
There is always an I-map $G \in \mathcal{G}$ of $\Prob \in \mathcal{P}^0$ which maximizes the CLL function \eqref{eq:CLL} on $\cD$ (if chain rule of probability applies, as we can use a full graph). Lemma \ref{lem:I-maps_2_CLL} tells us that all the I-maps $G$ of $\Prob$ maximize the CLL function. Lemma \ref{lem:I-map_in_G'} tells us that at least one of the I-maps belongs to $\mathcal{G}^1$. Hence, there is at least one I-map $G'\in \mathcal{G}^1$ which maximizes the CLL function \eqref{eq:CLL} on $\cD$. Or, equivalently, the relation \eqref{eq:Learning_problem_BND} holds.
\end{proof}

\subsection{Proposition \ref{pro:Decomposability}}

\begin{proof}
The fact that, for any $G \in \mathcal{G}^1$, the joint conditional distribution \eqref{eq:jointConditionalProbabilities} can be factorized as 
\begin{align*}
    \Prob^G_{\theta}(\by \given \bx) = \prod_{Y \in \mathbf{Y}} \Prob_{\theta} \left(y \given \pi_y\right) \,, \forall (\bx,\by) \in \cX \times \cY \,.
\end{align*}
can be checked easily.  Since $Y \notin \Delta^X$, for any $Y \in \mathbf{Y}$ and for any $X \in \mathbf{X}$, we have 
\begin{align}
     \Prob^G_\theta(\by \given \bx) &= \frac{\Prob^G_\theta(\bx,\by)}{\sum_{\by' \in \mathcal{Y}}\Prob^G_\theta(\bx,\by')} = \frac{\prod_{X\in \mathbf{X}} \Prob_\theta(x\given \pi_{x}) \prod_{Y\in \mathbf{Y}} \Prob_\theta(y\given \pi_{y})}{\sum_{\by'}\prod_{X\in \mathbf{X}} \Prob_\theta(x\given \pi_{x}) \prod_{Y\in \mathbf{Y}} \Prob_\theta(y'\given \pi_{y'})} \nonumber\\
    &= \frac{\prod_{X\in \mathbf{X}} \Prob_\theta(x\given \pi_{x}) \prod_{Y\in \mathbf{Y}} \Prob_\theta(y\given \pi_{y})}{\prod_{X\in \mathbf{X}} \Prob_\theta(x\given \pi_{x})\sum_{\by'} \prod_{Y\in \mathbf{Y}} \Prob_\theta(y'\given \pi_{y'})} = \frac{\prod_{Y\in \mathbf{Y}} \Prob_\theta(y\given \pi_{y})}{\sum_{\by'}\prod_{Y\in \mathbf{Y}} \Prob_\theta(y'\given \pi_{y'})} \label{eq:remove_feature_impact_11} \\
    &= \prod_{Y\in \mathbf{Y}} \Prob_\theta(y\given \pi_{y}) \,.\label{eq:remove_feature_impact_12}
\end{align}
The transition from \eqref{eq:remove_feature_impact_11} to \eqref{eq:remove_feature_impact_12} is straightforward because, by the definition of BNs, we have 
\begin{align}\label{eq:variable_elimination}
 \sum_{\by'}\prod_{Y\in \mathbf{Y}} \Prob_\theta(y'\given \pi_{y'}) = 1\,.
\end{align}

We now prove that following relation holds:
\begin{align*}
 \max_{(G,\theta) \in \mathcal{P}^1} C(\Prob^G_\theta\given \cD) = \max_{(G,\theta) \in \mathcal{P}^2} C(\Prob^G_\theta\given \cD) \,,
\end{align*}
We first partition $\mathcal{G}^1$ into $R(K) 2^{KQ}$ groups where each group consists of $R(Q)$ DAGs whose edges among $\mathbf{Y}$ and edges from features to class variables are the same. The relation 
\begin{align*}
    \Prob^G_\theta(\by \given \bx) = \prod_{Y \in \mathbf{Y}} \Prob_\theta \left(y \given \pi_y\right) \,, \forall (\bx,\by) \in \cX \times \cY 
\end{align*}  
ensures that all the members of each group have the same CLL score. Moreover, each group contains exactly one member of $\mathcal{G}^2$, i.e., the DAG with no edge among features. Therefore, the maximal CLL score attained over $\mathcal{G}^1$ equals the maximal score attained over $\mathcal{G}^2$.       
\end{proof}

\subsection{Proposition \ref{pro:Learning_problem_P3}}
Proof of Proposition \ref{pro:Learning_problem_P3} is trivial and is written down for completeness.
\begin{proof}
    For any $G \in \mathcal{P}^2$, there is at least one I-map $G' \in \mathcal{P}^3$ (to see that, simply add the extra arcs to $G$ to complete the parent sets of any class variable with all the continuous feature variables, leading to a graph $G' \in \mathcal{P}^3$ -- adding arcs will keep the I-map property). Thus, Proposition \ref{pro:Learning_problem_P3} comes as a consequence of Proposition \ref{pro:Decomposability}. 
\end{proof}

\subsection{Corollary \ref{cor:optimality}}

We would like to re-emphasize that our assumptions of having the optimality of learned parameters in the local models are not too strong. These are much weaker assumptions than those one finds in the literature when investigating the optimality of PGM learning frameworks: that is typically the assumption that the hypothesis space contains the possible distributions from some given family and the best estimate(s) converge to the optimal distribution(s) asymptotically. 

We do not require any asymptotic results, and the requirement of optimally learned parameters (given data) can be met by many standard estimation methods. Yet, this cannot be always guaranteed in practice, in particular if someone decides to use complicated models connecting the input feature variables and class variables, so our assumption is necessary for the proof of global optimality of the framework (which is a strong result and obviously cannot be achieved if base local models are not optimal themselves). 

With this in mind, the following (short) proof would satisfactorily inform readers of the significance of the proposed framework regarding the optimality.  

\begin{proof}
Assume the chain rule of probability holds (which is arguably a mild assumption) and the parameter learning problem is optimally solved. As a combination of Proposition \ref{pro:Learning_problem_BND}, Proposition \ref{pro:Decomposability} and Proposition \ref{pro:Learning_problem_P3}, we have 
\begin{align*}
\max_{\Prob\in\mathcal{P}^0} C(\Prob \given \cD) =\!\!\! \max_{(G, \theta) \in \mathcal{P}} C(\Prob^G_\theta \given \cD) =\!\!\! \max_{(G, \theta) \in \mathcal{P}^1} C(\Prob^G_\theta \given \cD) =\!\!\! \max_{(G, \theta) \in \mathcal{P}^2} C(\Prob^G_\theta \given \cD) =\!\!\! \max_{(G, \theta) \in \mathcal{P}^3} C(\Prob^G_\theta \given \cD)\,.
\end{align*}

Thus, algorithm \ref{alg:learn_GBNC} should return an I-map of the optimal distribution in $\mathcal{P}^0$. In other words, the learning procedure is universal, as $(G^*, \theta^*)$ is optimal with respect to $\mathcal{P}$, and with enough data would match the true conditional $\Prob(\mathbf{Y}|\mathbf{X})$ for $\Prob$ in $\mathcal{P}^0$.
\end{proof}

\subsection{Proposition \ref{pro:monotonicity}}
Enlarging parent sets (with discrete features) in our setting is analogous to further partitioning the input space in local supervised learning parts~\citep{wang2012local}. A representative of such approaches is the top-down construction of decision trees \citep{landwehr2005logistic,rokach2005top}. In such approaches, it is well-known that further partitioning the input space leads to higher predictive performance on the training data sets~\citep{landwehr2005logistic,rokach2005top}, as long as they are optimally learned. We can expect a similar phenomenon in our setting because CLL \eqref{eq:CLL} is indeed a performance measure for our probabilistic classifiers, and the way we encode each local distribution using $|\Pi^Y_d|$ distributions $\Prob_{\theta}\left(Y \given \pi, \mathbf{X}_c \right)$, $\forall \pi \in \Pi^Y_d$, makes our approach an input space partitioning approach, where $\pi \in \Pi^Y_d$ are used to partition the space formed by $\mathbf{X}_c$. 

We now present a proof for Proposition \ref{pro:monotonicity}.

\begin{proof}
Let $\Prob_{Y,\pi}$ be the local models used for $Y$ with parent set $\Delta$, for $\pi\in\Pi^Y_d$, and $\Prob'_{Y,\pi'}$ be the local models for parent set $\Delta'\supset\Delta$ such that $\pi\subset\pi'$. Let $\theta$ be the optimal parameters used by $\Prob_{Y,\pi}$. Because the local models remain as models from continuous features
$\mathbf{X}_c$ to the class variable $Y$, $\theta$ is still a valid solution (albeit non optimal) of the parameter learning of $\Prob'_{Y,\pi'}$, for each $\pi'$ extending $\pi$. If we use such a $\theta$ and sum together the CLL of all $\Prob'_{Y,\pi'}$ with $\pi'\supset\pi$ (that is, the extended parent set configurations that are compatible with $\pi$ over the variables they have in common), then we achieve the very same score~\eqref{eq:CLL}. Repeating this for all extension of all $\pi\in\Pi^Y_d$, the same overall CLL score is reached. This means that the CLL obtained after the added parents in $\Delta'\setminus\Delta$ has to be equal or larger (as it is assumed to be optimally learned) than before adding the parents. This proves Inequality \eqref{eq:monotonicity}. Now, \eqref{eq:monotonicity} guarantees that enlarging parent sets cannot decrease the CLL score \eqref{eq:CLL}. This ensures that at least one solution of the Algorithm \ref{alg:learn_GBNC} is a fully connected BN in the sense that the DAG over the class variables induced by its structure $G$ is fully connected. Such a solution can be found by finding a topological order of a solution $G$ from Algorithm~\ref{alg:learn_GBNC} and then adding arcs to $G$ (respecting that topological order) until the DAG over the class variables induced by the structure $G$ is fully connected (side comment: obviously it is not our goal to have fully connected networks, this is just to proof the theoretical results).
\end{proof}

\section{Detailed Algorithms}\label{sec:Algorithms_appendix}

\subsection{Algorithm \ref{alg:learn_GBNC}}\label{sec:Naive_Algorithm_appendix}

In this section, we show that Algorithm \ref{alg:learn_GBNC} can be revised to find a GBNC $(G^*,\theta^*) \in \mathcal{P}^3$ of any regularized variant \eqref{eq:regularized_CLL} of the CLL function. We first compute $ C(\Prob_{\theta^*_{Y,\pi}}\given Y,\pi,\cD)$ from $\mathcal{D}$ by solving \eqref{eq:P1} in line 5 of Algorithm \ref{alg:learn_GBNC}, which can be done by extracting the data set
\begin{align}
    \mathcal{D}_\pi \defi \left\{ (\bx_n,\by_n) \in \mathcal{D}\given \pi_{y_n } = \pi \right\} 
\end{align}
and calling any base learner to learn the optimal parameter set of $\Prob_{\theta}\left(Y \given \pi, \mathbf{X}_c\right)$ on $\mathcal{D}_\pi$ with respect to \eqref{eq:P1}.

For any given regularized variant \eqref{eq:regularized_CLL} of the CLL function, we denote by 
\begin{align}\label{eq:local_score_regularized_variants}
    S(Y, \Delta^Y_d)= C(Y, \Delta^Y_d) -  \text{pen}(\Delta^Y_d, |\mathcal{D}|) \,.
\end{align}
Clearly, the problem of learning a best $G \in \mathcal{G}^3$ can be re-expressed as an Integer Programming (IP):
\begin{align}
    \maximize  & \sum_{Y\in \mathbf{Y}}  \sum_{\Delta^Y_d\in \mathcal{F}^Y}  \gamma(\Delta^Y_d)  \cdot S(Y, \Delta^Y_d)   \,, \label{eq:regularized_target_function_GOBNILP}\\
     \text{Subject to }& 
     \sum_{\Delta^Y_d\in \mathcal{F}^Y} \gamma(\Delta^Y_d) =1 \,, \forall Y \in \mathbf{Y}  \, ,\nonumber \\
     &\sum_{Y \in \mathbf{Y}'} \sum_{\substack{\Delta^Y_d \in \mathcal{F}^Y \\ \Delta^Y_d \cap \mathbf{Y}' = \emptyset}} \gamma(\Delta^Y_d) >1  \,, \forall \mathbf{Y}' \subseteq \mathbf{Y}\,, |\mathbf{Y}'| >1 \, , \nonumber\\
     &\gamma(\Delta^Y_d)  \in \{0,1\} \,, \forall Y \in \mathbf{Y}, \forall ,\Delta^Y_d \in \mathcal{F}^Y \,.\nonumber
\end{align}  
Altogether, we end up with the implementation given in Algorithm \ref{alg:learn_GBNC_for_regularized_CLL}, which returns a GBNC $(G^*, \theta^*) \in \mathcal{P}^3$ of \eqref{eq:Learning_problem_BND_relaxed}.
   
	\begin{algorithm} [!ht]
	\caption{Learning a GBNC of \eqref{eq:Learning_problem_BND_relaxed} under the presence of regularization}\label{alg:learn_GBNC_for_regularized_CLL}
	\begin{algorithmic}[1]
   \STATE {\bfseries Input:} Data $\mathcal{D}$, Probabilistic hypothesis spaces encoding $\Prob_{\theta}\left(Y \given \pi, \mathbf{X}_c \right)$, $\forall \pi \in \Pi^Y_d$, $\forall \Delta^Y_d \in \mathcal{F}^Y$,  $\forall Y \in \mathbf{Y}$ \;
   \FOR{$Y \in \mathbf{Y}$}
        \FOR{$\Delta^Y_d \in \mathcal{F}^Y$}
            \FOR{$\pi \in \Pi^Y_d$}
                \STATE Solve \eqref{eq:P1} and store it in a proper data structure
            \ENDFOR
            \STATE Compute $S(Y,\Delta^Y_d)$ by \eqref{eq:local_score_regularized_variants} using stored values\;
        \ENDFOR
   \ENDFOR
 \STATE Find a best collection $\{\Delta^Y_d\given Y \in \mathbf{Y}\}$ which optimizes \eqref{eq:regularized_target_function_GOBNILP} using GOBNILP \;
   \STATE {\bfseries Output:} A GBNC $(G^*, \theta^*) \in \mathcal{P}^3$ of \eqref{eq:Learning_problem_BND_relaxed} \;
   \end{algorithmic}
   \end{algorithm}

\subsection{A Refinement of Algorithm \ref{alg:learn_GBNC}}\label{sec:Refine_Algorithm_appendix}

In this section, we show how pruning rules \citep{de2018entropy} can be employed to find GBNCs which optimize regularized variants \eqref{eq:regularized_CLL} without losing any optimality. 

We first generalize the pruning rule \citep{de2018entropy}[Lemma $3$] for any regularized variant of the form \eqref{eq:regularized_CLL}. 
\begin{lemma}\label{eq:lem_regularized_CLL}
Let $Y\in \mathbf{Y}$ be a node in $G \in \mathcal{G}^3$ with $\Delta \subset \Delta' \in \mathcal{F}^Y$, such that 
\begin{align}\label{eq:pruning_rule}
S(Y,\Delta) \geq - \text{pen}(\Delta', |\mathcal{D}|) \,.   
\end{align}
Then $\Delta'$ and all its supersets can be safely discarded from $\mathcal{F}^Y$ without decreasing the maximum score of \eqref{eq:Learning_problem_BND_relaxed}.
\end{lemma}
\begin{proof}
Using the shorthand notation \eqref{eq:local_score_regularized_variants}, a regularized variant of the CLL function can be rewritten as 
\begin{equation*}
S(\Prob^G_{\theta^*} \given \mathcal{D}) =  \sum_{Y \in \mathbf{Y}}  S(Y,\Delta^Y_d) =  \sum_{Y \in \mathbf{Y}} \left( C(Y,\Delta^Y_d)  -  \text{pen}(\Delta^Y_d, |\mathcal{D}|) \right)\,.
\end{equation*}
For any $G, G' \in \mathcal{G}^3$ such that $\Delta$ and $\Delta'$ are respectively the parent set of $Y$ in $G$ and $G'$, and the parent sets of all $Y'\neq Y$ are the same, we have the relation 
\begin{align*}
 S(\Prob^G_{\theta^*} \given \mathcal{D}) -  S(\Prob^{G'}_{\theta^*} \given \mathcal{D}) &=     S(Y,\Delta) -  S(Y,\Delta') = S(Y,\Delta)- C(Y,\Delta) +  \text{pen}(\Delta', |\mathcal{D}|) \\
 &\geq -  \text{pen}(\Delta', |\mathcal{D}|) -  C(Y,\Delta) +  \text{pen}(\Delta', |\mathcal{D}|) = -  C(Y,\Delta) \geq 0\,.
\end{align*}
Thus, we can safely discard  $\Delta'$ from $\mathcal{F}^Y$ without decreasing the maximum score of \eqref{eq:Learning_problem_BND_relaxed}. 

Because for any $\Delta' \subset \Delta'' \in \mathcal{F}^Y$, we have $ -  \text{pen}(\Delta', |\mathcal{D}|)  \geq   -  \text{pen}(\Delta'', |\mathcal{D}|)$ and assumption \eqref{eq:monotonicity} ensures that $C(Y,\Delta) \leq C(Y,\Delta'')$. Thus, we have the relation 
\begin{align*}
S(Y,\Delta) \geq - \text{pen}(\Delta', |\mathcal{D}|)  \geq   -  \text{pen}(\Delta'', |\mathcal{D}|)  \,.   
\end{align*}
which ensures that we can also safely discard $\Delta''$ from $\mathcal{F}^Y$. 
\end{proof}
Intuitively, Lemma \ref{eq:lem_regularized_CLL} provides us a "stopping criterion" for enlarging parent sets by exploiting the fact that regularized variants \eqref{eq:regularized_CLL} of the CLL \eqref{eq:CLL} seek for a trade-off between the predictive performance provided by more complex classifiers and the simplicity of classifiers. More precisely, condition \eqref{eq:pruning_rule} allows one to safely discard (some/many large) possible parent set $\Delta'$ and its supersets $\Delta''$ without the need of learning local probabilistic classifiers \eqref{eq:Local_Classifiers} for these parent sets. It is very beneficial, because if we do not have such a stopping criterion, we will need to evaluate all the possible parent sets and evaluating each $\Delta^Y \in \mathcal{F}^Y$ requires one to the learning a possibly large number of local probabilistic classifiers \eqref{eq:Local_Classifiers} which is exponential in the cardinality of $\Delta^Y$.   

Ideally, we would expect that enlarging the parent sets (or increasing the model complexity) gives us a better score $S(Y,\Delta^Y)$, i.e., assumption \eqref{eq:monotonicity} should hold. However, in practice, it may happen that the learning algorithm fails to converge and returns unreliable (and inaccurate) local probabilistic classifiers \eqref{eq:Local_Classifiers}. In such a case, we would keep adding redundant parents and end up with unreliable local probabilistic classifiers \eqref{eq:Local_Classifiers} in the final GBNC. In other words, we pick up an unnecessary complex GBNC which contains unreliable local probabilistic classifiers \eqref{eq:Local_Classifiers}. To avoid this unexpected behavior, we propose a variant of the pruning rule \eqref{eq:pruning_rule}.  

\begin{definition}\label{def:pruning_rule_1}
Let $Y\in \mathbf{Y}$ be a node in $G \in \mathcal{G}^3$ with $\Delta \subset \Delta' \in \mathcal{F}^Y$, such that 
\begin{align}\label{eq:pruning_rule_1}
\max_{\Delta'' \subset \Delta} S(Y, \Delta'') \geq -  \text{pen}(\Delta', |\mathcal{D}|)  \,.   
\end{align}
Then all the $\Delta \supset \Delta^*$, where 
\begin{align}\label{eq:best_subset}
\Delta^* = \argmax_{\Delta'' \subset \Delta} S(Y, \Delta'') \,,    
\end{align}
will be discarded from $\mathcal{F}^Y$.
\end{definition}
Intuitively, the pruning rule \eqref{eq:pruning_rule_1} --\eqref{eq:best_subset} allows us to prune all the supersets of $\Delta^*$. For example, if $\Delta^* \subsetneq \Delta$, we discard all of its supersets, such as $\Delta$, $\Delta'$ and their supersets. Adopting the pruning rule \eqref{eq:pruning_rule_1} --\eqref{eq:best_subset}, we propose a refinement of Algorithm \ref{alg:learn_GBNC} which is summarized in Algorithm \ref{alg:learn_GBNC_refined}. To simplify Pseudocode, for any $Y \in \mathbf{Y}$, we denote by 
\begin{align}
    \mathcal{F}^Y_k = \left\{\Delta \in \mathcal{F}^Y \given |\Delta| = k \right\}\,, \forall k = |\mathbf{X}_c|, \ldots, Q + K \,.
\end{align}
Algorithm \ref{alg:learn_GBNC_refined} only learns a local classifier which estimates the local distributions $\Prob_{\theta}\left(Y \given \pi , \mathbf{X}_c\right)$, $\pi \in \Pi$, if $\Delta$ is still included in $\mathcal{F}^Y$ and its complexity is not so high according to \eqref{eq:pruning_rule_1}. In practice, we observed that large $\Delta \in \mathcal{F}^Y$ are usually discarded.  

	\begin{algorithm} [!ht]
	\caption{Learning a GBNC of \eqref{eq:Learning_problem_BND_relaxed} under the presence of regularization}\label{alg:learn_GBNC_refined}
	\begin{algorithmic}[1]
   \STATE {\bfseries Input:} Data $\mathcal{D}$, Probabilistic hypothesis spaces encoding $\Prob_{\theta}\left(Y \given \pi, \mathbf{X}_c \right)$, $\forall \pi \in \Pi^Y_d$, $\forall \Delta^Y_d \in \mathcal{F}^Y$,  $\forall Y \in \mathbf{Y}$  \;
   \FOR{$Y \in \mathbf{Y}$}
        \FOR{$k = |\mathbf{Y}_c|, \ldots, Q + K$}
            \FOR{$\Delta^Y_d \in \mathcal{F}^Y_k$}
            \IF{Condition \eqref{eq:pruning_rule_1} holds}
            \STATE Determine $\Delta^*$ using \eqref{eq:best_subset}; Discard all the $\Delta^Y_d \supset \Delta^*$ from $\mathcal{F}^Y$ \;
            \ELSE
            \STATE  Compute $S(Y,\Delta^Y_d)$ defined in \eqref{eq:local_score_regularized_variants}; Store $\Prob_{\theta^*}\left(Y \given \pi, \mathbf{X}_c\right)$, $\forall \pi \in \Pi^Y_d$ in a proper data structure \;
            \ENDIF
            \ENDFOR
        \ENDFOR
   \ENDFOR
   \STATE Find a best collection $\{\Delta^Y_d\given Y \in \mathbf{Y}\}$ which optimizes \eqref{eq:regularized_target_function_GOBNILP} using GOBNILP \;
   \STATE {\bfseries Output:} A GBNC $(G, \theta) \in \mathcal{P}^3$ of \eqref{eq:Learning_problem_BND_relaxed} \;
   \end{algorithmic}
   \end{algorithm}

\subsection{Inference Algorithms}

Practical procedures for finding BOPs of $\ell_S$ \eqref{eq:BOP_Subset} and $\ell_H$ \eqref{eq:BOP_Hamming} are presented in Algorithm \ref{alg:BOP_Subset} and Algorithm \ref{alg:BOP_Hamming}, respectively.

	\begin{algorithm} [!ht]
	\caption{Find a BOP of the Subset $0/1$ loss \eqref{eq:BOP_Subset}}\label{alg:BOP_Subset}
	\begin{algorithmic}[1]
    \STATE {\bfseries Input:} A GBNC $(G^*, \theta^*) \in \mathcal{P}^3$ of \eqref{eq:Learning_problem_BND_relaxed}: $\Prob_{\theta^*}\left(Y \given \pi, \mathbf{X}_c \right)$, $\forall \pi \in \Pi^Y_d$, $\forall \Delta^Y_d \in \mathcal{F}^Y$,  $\forall Y \in \mathbf{Y}$, a test instance $\bx$ \;
    \STATE Extract the sub-DAG $\mathcal{K}$ over $\mathbf{Y}$ from $G$\;
    \FOR{$Y \in \mathbf{Y}$}
    \STATE Extract parent set of $Y$ in $\mathcal{K}$: $\Delta^Y_\mathbf{Y} = \Delta^Y_d \cap \mathbf{Y}$; Form the set $\Pi^Y_\mathbf{Y}$ of the possible configurations of $\Delta^Y_\mathbf{Y}$\;
        \FOR{$\pi^Y_\mathbf{Y} \in \Pi^Y_\mathbf{Y}$}
            \STATE Predict $\Prob^{\mathcal{K}}_{\theta^*}\left(Y \given \pi^Y_\mathbf{Y}\right)$ using $\Prob_{\theta^*}\left(Y \given \pi, \mathbf{X}_c \right)$ which are specified by $\bx_d$\;
        \ENDFOR
   \ENDFOR
   \STATE Find a MPE $\hat{\by} \in \mathcal{Y}$ given $\mathcal{K}$ and $\Prob^{\mathcal{K}}_{\theta^*}\left(Y \given \pi^Y_\mathbf{Y}\right)$, $\forall \pi^Y_\mathbf{Y} \in \Pi^Y_\mathbf{Y}$, $\forall Y \in \mathbf{Y}$\;
   \STATE {\bfseries Output:} A BOP $\hat{\by}$ of the Subset $0/1$ loss \eqref{eq:BOP_Subset} \;
   \end{algorithmic}
   \end{algorithm}

   	\begin{algorithm} [!ht]
	\caption{Find a BOP of the Hamming loss \eqref{eq:BOP_Hamming}}\label{alg:BOP_Hamming}
	\begin{algorithmic}[1]
    \STATE {\bfseries Input:} A GBNC $(G^*, \theta^*) \in \mathcal{P}^3$ of \eqref{eq:Learning_problem_BND_relaxed}: $\Prob_{\theta^*}\left(Y \given \pi, \mathbf{X}_c \right)$, $\forall \pi \in \Pi^Y_d$, $\forall \Delta^Y_d \in \mathcal{F}^Y$,  $\forall Y \in \mathbf{Y}$, a test instance $\bx$ \;
    \STATE Extract the sub-DAG $\mathcal{K}$ over $\mathbf{Y}$ from $G$\;
    \FOR{$Y \in \mathbf{Y}$}
    \STATE Extract parent set of $Y$ in $\mathcal{K}$: $\Delta^Y_\mathbf{Y} = \Delta^Y_d \cap \mathbf{Y}$; Form the set $\Pi^Y_\mathbf{Y}$ of the possible configurations of $\Delta^Y_\mathbf{Y}$\;
        \FOR{$\pi^Y_\mathbf{Y} \in \Pi^Y_\mathbf{Y}$}
            \STATE Predict $\Prob^{\mathcal{K}}_{\theta^*}\left(Y \given \pi^Y_\mathbf{Y}\right)$ using $\Prob_{\theta^*}\left(Y \given \pi, \mathbf{X}_c \right)$ which are specified by $\bx_d$\;
        \ENDFOR
   \ENDFOR
   \STATE Find $K$ marginals $\hat{y}^1, \ldots, \hat{y}^K$ given $\mathcal{K}$ and $\Prob^{\mathcal{K}}_{\theta^*}\left(Y \given \pi^Y_\mathbf{Y}\right)$, $\forall \pi^Y_\mathbf{Y} \in \Pi^Y_\mathbf{Y}$, $\forall Y \in \mathbf{Y}$\;
   \STATE {\bfseries Output:} A BOP $\hat{\by} \defi (\hat{y}^1, \ldots, \hat{y}^K)$ of the Hamming loss \eqref{eq:BOP_Hamming} \;
   \end{algorithmic}
   \end{algorithm}

\section{The Case of Partial/Missing Data}\label{sec:Missing_data}

The structural Expectation-Maximization (structural EM) approach has been used in different works in BN learning from missing data \citep{adel2017learning,rancoita2016,friedman1998bayesian}. Reminding that, in BN learning with an incomplete training data, the structural EM approach \citep{friedman1998bayesian} can be employed to find a pair of a possible precise/complete data set and a possible BN, which optimizes some given target function. 

The structural EM approach can be implemented as a two-step algorithm, which should be iterated until either the algorithm converges or some stopping criterion is met. 

\begin{itemize}
    \item Expectation step (E): we complete the data by imputing partial/missing data from a fitted BN;
    \item Maximization step (M): we learn a BN by optimizing given target function over the completed data.
\end{itemize}

Yet, we can in principle adapt the M step of the structural EM approach to the setting of probabilistic MDC straightforwardly. 

However, depending on the concrete type of missing data we are dealing with, handling the E step may require more attention. In the case of partially specified class variables and precise features \citep{wang2021learning}, GBNCs given by Algorithm \ref{alg:learn_GBNC} and \ref{alg:learn_GBNC_for_regularized_CLL} are estimates of $\Prob(\cY \vert \cX)$ and can be used to impute partial/missing data during the E step. In the general case where both the features and class variables can be partially specified \citep{hullermeier2014learning,nguyen2021racing}, estimates of $\Prob(\cY \vert \cX)$ itself seems to be inadequate, because estimates of $\Prob(\cX, \cY)$ may be needed for doing imputation if one wishes to use exact/approximate inference. We however leave this problem as a future work because it is beyond the scope of this paper.    

\section{Experiments}\label{sec:Experiments_appendix}

\subsection{Experimental Setting}\label{sec:Experimental_Setting_appendix}

We evaluate our approaches on both tabular and image data sets. \autoref{tab:stats_tab_data} summarizes the detailed statistics of all tabular data sets, which are originally collected by \citep{jia2021decomposition}. 
From left to right, the meaning of each column is the number of class variables (\#CV), the number of samples (\#Samples), and the number of states of each class variable. (\#States/CV) and the number of features (\#Features), respectively.  Among the 20 tabular data sets, there are three data sets (Adult, Default and Thyroid) which contain mixed features. If all class variables contain the same number of states, only this number is reported. For example, the Flickr data set has five class variables, which have 3, 4, 3, 4, and 4 states, respectively. 

\begin{table}[!ht]
  \centering
\caption{Statistics of the tabular benchmark data sets.}
\label{tab:stats_tab_data}
\begin{tabular}{lcccc}
\toprule 
Data Set     & \#CV & \#States/CV             & \#Samples & \#Features\\ \hline
Edm          & 2    & 3                        & 154      & $16 n$    \\
Jura         & 2    & 4,5                      & 359      & $9 n$     \\
Enb          & 2    & 2,4                      & 768      & $6 n$     \\
Voice        & 2    & 4,2                      & 3136     & $19 n$    \\
Song         & 3    & 3                        & 785      & $98 n$    \\
Adult        & 4    & $7,7,5,2$                & 18419    & $5 n, 5 x$  \\
Default      & 4    & $2,7,4,2$                & 28779    & $14 n, 6 x$ \\ 
Flickr       & 5    & $3,4,3,4,4$              & 12198    & $1536 n$  \\
Fera         & 5    & 6                        & 14052    & $136 n$   \\
\hline
WQplants     & 7    & 4                        & 1060     & $16 n$    \\
WQanimals    & 7    & 4                        & 1060     & $16 n$    \\
Thyroid      & 7    & $5,5,3,2,4,4,3$          & 9172     & $7 n, 22 x$ \\
\hline
Rf1          & 8    & $4,4,3,4,4,3,4,3$        & 8987     & $64 n$    \\
Pain         & 10   & $2,5,4,2,2,5,2,5,2,2$    & 9734     & $136 n$   \\
Disfa        & 12   & $5,5,6,3,4,4,5,4,4,4,6,4$& 13095    & $136 n$   \\
WaterQuality & 14   & 4                        & 1060     & $16 n$    \\
Oes97        & 16   & 3                        & 334      & $263 n$   \\
Oes10        & 16   & 3                        & 403      & $298 n$   \\
Scm20d       & 16   & 4                        & 8966     & $61 n$    \\
Scm1d        & 16   & 4                        & 9803     & $280 n$   \\
\bottomrule
\end{tabular}
\end{table}

We compare two instantiations of \GBNCs{} (\GBNC-S which optimizes \eqref{eq:regularized_CLL} and produces BOP \eqref{eq:BOP_Subset} of $\ell_S$, and \GBNC-H which optimizes \eqref{eq:regularized_CLL} and produces BOP \eqref{eq:BOP_Hamming} of $\ell_H$) with BR, CP, and CCs \citep{jia2021decomposition}[Section II--III]. Reminding that $\text{pen}(\Delta^Y_d, |\mathcal{D}|)$ is the penalty term of the Bayesian Information Criterion
(BIC) \citep{schwarz1978estimating} in our experiments. 

It is known that the chain order of CCs can (significantly) affect its performance and choosing the best order is one of the toughest problems in learning CCs \citep{read2021classifier}. Although choosing good orders of CCs is not a focus of our work, randomly choosing orders would make CCs too weak. We thus sample 11 orders (which are the original order of the class variables from the data source and 10 other orders generated randomly) and pick the best chain order in terms of validation performance, i.e., we use 80\% of the training data to learn CCs and pick up the most promising order with the highest performance on the validation set consists of 20\% of training data, and report its test performance. When running the experiments, we observed that this often improves the performance of CCs, compared to randomly choosing one chain order. We follow the suggestion of \citep{jia2021decomposition} and convert discrete features/variables into continuous variables using one-hot encoding whenever they appear as parts of input of local classifiers of BR, CP and CCs. While there are other multi-dimensional classifiers with promising predictive performances, such as \citep{jia2021decomposition} and \citep{jia2023multi}, we find it hard to interpret such classifiers as probabilistic classifiers. Thus, we do not include them in our experimental comparison, which specifically focuses on probabilistic classifiers. 

For each tabular data set, we do a 10-fold cross-validation, and report the mean and standard deviation of the performance of the classifiers. For the image data set, we do a 3-fold cross-validation, and report the mean and standard deviation of the performance of the classifiers. 

We implement all approaches in Python and use the pgmpy framework \citep{ankan2015pgmpy}. 
We use the PyTorch framework \citep{paszke2019pytorch} to implement neural networks. The source code to replicate experiments is provided as supplementary materials and has been made made public at \url{https://github.com/yangyang-pro/probabilistic-mdc}.

\subsection{Results}\label{sec:Results_appendix}

This appendix provides detailed experimental results which are summarized in section \ref{sec:Experiments} of the main text.

\begin{figure}[!ht]
\centering
\begin{tabular}{cc}
\begin{tikzpicture}[scale = 0.8]
\begin{axis}[%
scatter/classes={scatter src = explicit symbolic,%
    w={mark=*,draw=black},
    b={mark=*,draw=black},
    c={mark=*,draw=black},
    d={mark=*,draw=black},
    e={mark=*,draw=black},
    f={mark=*,draw=black},
    g={mark=*,draw=black},
    h={mark=*,draw=black},
    i={mark=*,draw=black},
    j={mark=*,draw=black},
    k={mark=*,draw=black},
    l={mark=*,draw=black},
    m={mark=*,draw=black},
    n={mark=*,draw=black},
    o={mark=*,draw=black},
    p={mark=*,draw=black},
    q={mark=*,draw=black},
    r={mark=*,draw=black},
    s={mark=*,draw=black},
    t={mark=*,draw=black}}, xlabel = GBNC-H,
ylabel = BR]
\addplot[draw=blue,pattern=horizontal lines light blue]
 coordinates { (0.02,0.02) (0.4,0.4) };
\addplot[scatter,only marks,%
    scatter src=explicit symbolic]%
table[meta=label] {
x       y       label
0.2654	0.2765	w
0.3732	0.3551	b
0.2220	0.2416	c
0.0821	0.0808	d
0.2433	0.2574	e
0.2174	0.2022	f
0.3776	0.3882	g
0.3461	0.3465	h
0.3685	0.3698	i
0.0953	0.1610	j
0.0470	0.0474	k
0.1030	0.1058	l
0.3553	0.3610	m
0.2761	0.2824	n
0.1955	0.1921	o
0.3145	0.3604	p
0.1807	0.2342	q
0.3246	0.2826	r
0.3329	0.3348	s
0.0338	0.0352	t
    };
\end{axis}
\end{tikzpicture} 
&
\begin{tikzpicture}[scale = 0.8]
\centering
\begin{axis}[%
scatter/classes={scatter src = explicit symbolic,%
    w={mark=*,draw=black},
    b={mark=*,draw=black},
    c={mark=*,draw=black},
    d={mark=*,draw=black},
    e={mark=*,draw=black},
    f={mark=*,draw=black},
    g={mark=*,draw=black},
    h={mark=*,draw=black},
    i={mark=*,draw=black},
    j={mark=*,draw=black},
    k={mark=*,draw=black},
    l={mark=*,draw=black},
    m={mark=*,draw=black},
    n={mark=*,draw=black},
    o={mark=*,draw=black},
    p={mark=*,draw=black},
    q={mark=*,draw=black},
    r={mark=*,draw=black},
    s={mark=*,draw=black},
    t={mark=*,draw=black}}, xlabel = BR,
ylabel = CC]
\addplot[draw=blue,pattern=horizontal lines light blue]
 coordinates { (0.02,0.02) (0.42,0.42) };

\addplot[scatter,only marks,%
    scatter src=explicit symbolic]%
table[meta=label] {
x       y       label
0.2765	0.2723	w
0.3551	0.3981	b
0.2416	0.2220	c
0.0808	0.0808	d
0.2574	0.2646	e
0.2022	0.2046	f
0.3882	0.3923	g
0.3465	0.3542	h
0.3698	0.3823	i
0.1610	0.1633	j
0.0474	0.0491	k
0.1058	0.1063	l
0.3610	0.3650	m
0.2824	0.2935	n
0.1921	0.2080	o
0.3604	0.3815	p
0.2342	0.2563	q
0.2826	0.2846	r
0.3348	0.3339	s
0.0352	0.0344	t
    };
\end{axis}
\end{tikzpicture}
\\
\\
\\
\begin{tikzpicture}[scale = 0.8]
\begin{axis}[%
scatter/classes={scatter src = explicit symbolic,%
    w={mark=*,draw=black},
    b={mark=*,draw=black},
    c={mark=*,draw=black},
    d={mark=*,draw=black},
    e={mark=*,draw=black},
    f={mark=*,draw=black},
    g={mark=*,draw=black},
    h={mark=*,draw=black},
    i={mark=*,draw=black},
    j={mark=*,draw=black},
    k={mark=*,draw=black},
    l={mark=*,draw=black},
    m={mark=*,draw=black},
    n={mark=*,draw=black},
    o={mark=*,draw=black},
    p={mark=*,draw=black},
    q={mark=*,draw=black},
    r={mark=*,draw=black},
    s={mark=*,draw=black},
    t={mark=*,draw=black}}, xlabel = GBNC-H,
ylabel = CC]
\addplot[draw=blue,pattern=horizontal lines light blue]
 coordinates { (0.02,0.02) (0.4,0.4) };

\addplot[scatter,only marks,%
    scatter src=explicit symbolic]%
table[meta=label] {
x       y       label
0.2654	0.2723	w
0.3732	0.3981	b
0.2220	0.2220	c
0.0821	0.0808	d
0.2433	0.2646	e
0.2174	0.2046	f
0.3776	0.3923	g
0.3461	0.3542	h
0.3685	0.3823	i
0.0953	0.1633	j
0.0470	0.0491	k
0.1030	0.1063	l
0.3553	0.3650	m
0.2761	0.2935	n
0.1955	0.2080	o
0.3145	0.3815	p
0.1807	0.2563	q
0.3246	0.2846	r
0.3329	0.3339	s
0.0338	0.0344	t
    };
\end{axis}
\end{tikzpicture} 
&
\begin{tikzpicture}[scale = 0.8]
\begin{axis}[%
scatter/classes={scatter src = explicit symbolic,%
    w={mark=*,draw=black},
    b={mark=*,draw=black},
    c={mark=*,draw=black},
    d={mark=*,draw=black},
    e={mark=*,draw=black},
    f={mark=*,draw=black},
    g={mark=*,draw=black},
    h={mark=*,draw=black},
    i={mark=*,draw=black},
    j={mark=*,draw=black},
    k={mark=*,draw=black},
    l={mark=*,draw=black},
    m={mark=*,draw=black},
    n={mark=*,draw=black},
    o={mark=*,draw=black},
    p={mark=*,draw=black},
    q={mark=*,draw=black},
    r={mark=*,draw=black},
    s={mark=*,draw=black},
    t={mark=*,draw=black}}, xlabel = BR,
ylabel = CP]
\addplot[draw=blue,pattern=horizontal lines light blue]
 coordinates { (0,0) (0.7,0.7) };

\addplot[scatter,only marks,%
    scatter src=explicit symbolic]%
table[meta=label] {
x       y       label
0.2765	0.2823	w
0.3551	0.6784	b
0.2416	0.3177	c
0.0808	0.4169	d
0.2574	0.4930	e
0.2022	0.4921	f
0.3882	0.5297	g
0.3465	0.3806	h
0.3698	0.4307	i
0.1610	0.3649	j
0.0474	0.0524	k
0.1058	0.1308	l
0.3610	0.4092	m
0.2824	0.4559	n
0.1921	0.3840	o
0.3604	0.5787	p
0.2342	0.5559	q
0.2826	0.4140	r
0.3348	0.4394	s
0.0352	0.0389	t
    };
\end{axis}
\end{tikzpicture}
\\
\\
\\
\begin{tikzpicture}[scale = 0.8]
\begin{axis}[%
scatter/classes={scatter src = explicit symbolic,%
    w={mark=*,draw=black},
    b={mark=*,draw=black},
    c={mark=*,draw=black},
    d={mark=*,draw=black},
    e={mark=*,draw=black},
    f={mark=*,draw=black},
    g={mark=*,draw=black},
    h={mark=*,draw=black},
    i={mark=*,draw=black},
    j={mark=*,draw=black},
    k={mark=*,draw=black},
    l={mark=*,draw=black},
    m={mark=*,draw=black},
    n={mark=*,draw=black},
    o={mark=*,draw=black},
    p={mark=*,draw=black},
    q={mark=*,draw=black},
    r={mark=*,draw=black},
    s={mark=*,draw=black},
    t={mark=*,draw=black}}, xlabel = GBNC-H,
ylabel = CP]
\addplot[draw=blue,pattern=horizontal lines light blue]
 coordinates { (0,0) (0.7,0.7) };

\addplot[scatter,only marks,%
    scatter src=explicit symbolic]%
table[meta=label] {
x       y       label
0.2654	0.2823	w
0.3732	0.6784	b
0.2220	0.3177	c
0.0821	0.4169	d
0.2433	0.4930	e
0.2174	0.4921	f
0.3776	0.5297	g
0.3461	0.3806	h
0.3685	0.4307	i
0.0953	0.3649	j
0.0470	0.0524	k
0.1030	0.1308	l
0.3553	0.4092	m
0.2761	0.4559	n
0.1955	0.3840	n
0.3145	0.5787	p
0.1807	0.5559	q
0.3246	0.4140	r
0.3329	0.4394	s
0.0338	0.0389	t
    };
\end{axis}
\end{tikzpicture}
&
\begin{tikzpicture}[scale = 0.8]
\begin{axis}[%
scatter/classes={scatter src = explicit symbolic,%
    w={mark=*,draw=black},
    b={mark=*,draw=black},
    c={mark=*,draw=black},
    d={mark=*,draw=black},
    e={mark=*,draw=black},
    f={mark=*,draw=black},
    g={mark=*,draw=black},
    h={mark=*,draw=black},
    i={mark=*,draw=black},
    j={mark=*,draw=black},
    k={mark=*,draw=black},
    l={mark=*,draw=black},
    m={mark=*,draw=black},
    n={mark=*,draw=black},
    o={mark=*,draw=black},
    p={mark=*,draw=black},
    q={mark=*,draw=black},
    r={mark=*,draw=black},
    s={mark=*,draw=black},
    t={mark=*,draw=black}}, xlabel = CC,
ylabel = CP]
\addplot[draw=blue,pattern=horizontal lines light blue]
 coordinates { (0,0) (0.7,0.7) };

\addplot[scatter,only marks,%
    scatter src=explicit symbolic]%
table[meta=label] {
x       y       label
0.2723	0.2823	w
0.3981	0.6784	b
0.2220	0.3177	c
0.0808	0.4169	d
0.2646	0.4930	e
0.2046	0.4921	f
0.3923	0.5297	g
0.3542	0.3806	h
0.3823	0.4307	i
0.1633	0.3649	j
0.0491	0.0524	k
0.1063	0.1308	l
0.3650	0.4092	m
0.2935	0.4559	n
0.2080	0.3840	o
0.3815	0.5787	p
0.2563	0.5559	q
0.2846	0.4140	r
0.3339	0.4394	s
0.0344	0.0389	t
    };
\end{axis}
\end{tikzpicture}
\\
\end{tabular}
\caption{Hamming loss (\textbf{base learner: \textit{Logistic regression}})}
\label{fig:Hamming loss for pairs with LR}
\end{figure}
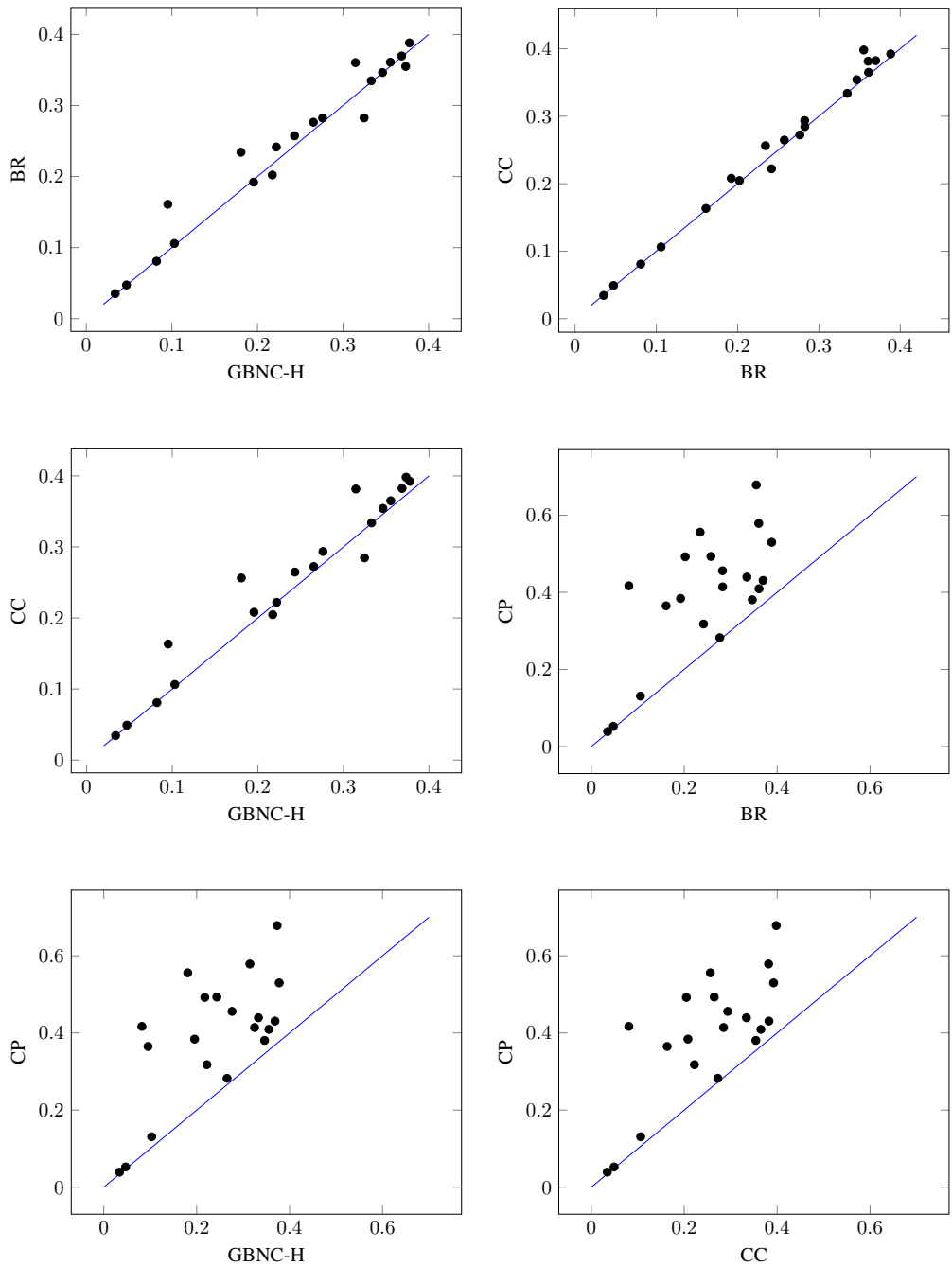

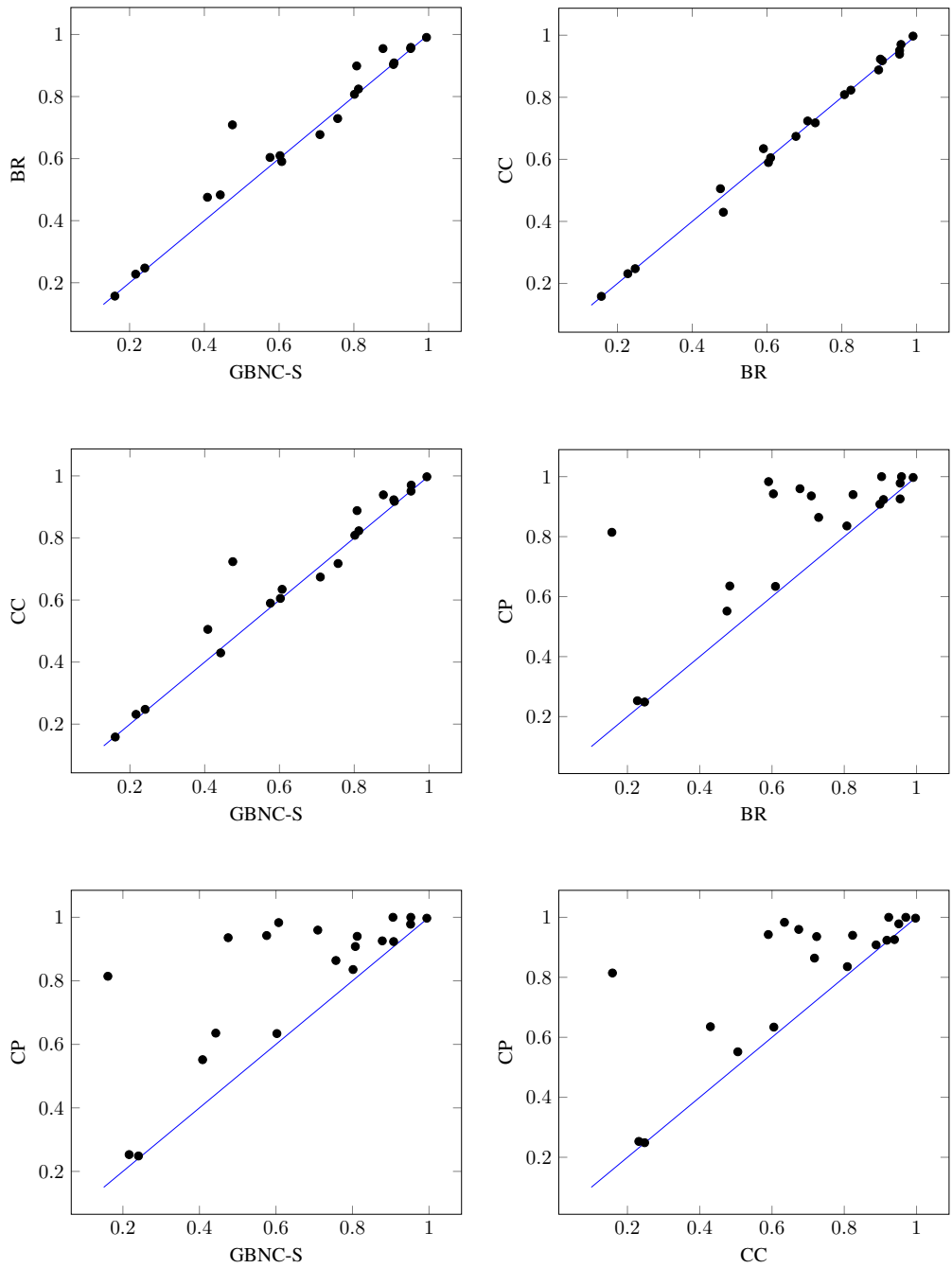
\begin{figure}[!ht]
\centering
\begin{tabular}{cc}
\begin{tikzpicture}[scale = 0.8]]
\begin{axis}[%
scatter/classes={scatter src = explicit symbolic,%
    w={mark=*,draw=black},
    b={mark=*,draw=black},
    c={mark=*,draw=black},
    d={mark=*,draw=black},
    e={mark=*,draw=black},
    f={mark=*,draw=black},
    g={mark=*,draw=black},
    h={mark=*,draw=black},
    i={mark=*,draw=black},
    j={mark=*,draw=black},
    k={mark=*,draw=black},
    l={mark=*,draw=black},
    m={mark=*,draw=black},
    n={mark=*,draw=black},
    o={mark=*,draw=black},
    p={mark=*,draw=black},
    q={mark=*,draw=black},
    r={mark=*,draw=black},
    s={mark=*,draw=black},
    t={mark=*,draw=black}}, xlabel = GBNC-S,
ylabel = BR]
\addplot[draw=blue,pattern=horizontal lines light blue]
 coordinates { (0.13,0.13) (1,1) };

\addplot[scatter,only marks,%
    scatter src=explicit symbolic]%
table[meta=label] {
x       y       label
0.4083	0.4754	w
0.6071	0.5905	b
0.4427	0.4832	c
0.1607	0.1569	d
0.5757	0.6038	e
0.7093	0.6774	f
0.8015	0.8072	g
0.9075	0.9085	h
0.9519	0.9547	i
0.4751	0.7088	j
0.2407	0.2474	k
0.6024	0.6096	l
0.9943	0.9906	m
0.9524	0.9584	n
0.9058	0.9033	o
0.8779	0.9546	p
0.8075	0.8986	q
0.7567	0.7291	r
0.8123	0.8243	s
0.2163	0.2276	t
    };
\end{axis}
\end{tikzpicture} 
&
\begin{tikzpicture}[scale = 0.8]]
\begin{axis}[%
scatter/classes={scatter src = explicit symbolic,%
    w={mark=*,draw=black},
    b={mark=*,draw=black},
    c={mark=*,draw=black},
    d={mark=*,draw=black},
    e={mark=*,draw=black},
    f={mark=*,draw=black},
    g={mark=*,draw=black},
    h={mark=*,draw=black},
    i={mark=*,draw=black},
    j={mark=*,draw=black},
    k={mark=*,draw=black},
    l={mark=*,draw=black},
    m={mark=*,draw=black},
    n={mark=*,draw=black},
    o={mark=*,draw=black},
    p={mark=*,draw=black},
    q={mark=*,draw=black},
    r={mark=*,draw=black},
    s={mark=*,draw=black},
    t={mark=*,draw=black}}, xlabel = BR,
ylabel = CC]
\addplot[draw=blue,pattern=horizontal lines light blue]
 coordinates { (0.13,0.13) (1,1) };

\addplot[scatter,only marks,%
    scatter src=explicit symbolic]%
table[meta=label] {
x       y       label
0.4754	0.5054	w
0.5905	0.6345	b
0.4832	0.4296	c
0.1569	0.1585	d
0.6038	0.5898	e
0.6774	0.6740	f
0.8072	0.8086	g
0.9085	0.9179	h
0.9547	0.9509	i
0.7088	0.7236	j
0.2474	0.2476	k
0.6096	0.6051	l
0.9906	0.9972	m
0.9584	0.9703	n
0.9033	0.9230	o
0.9546	0.9389	p
0.8986	0.8881	q
0.7291	0.7176	r
0.8243	0.8231	s
0.2276	0.2315	t
    };
\end{axis}
\end{tikzpicture}
\\
\\
\\
\begin{tikzpicture}[scale = 0.8]]
\begin{axis}[%
scatter/classes={scatter src = explicit symbolic,%
    w={mark=*,draw=black},
    b={mark=*,draw=black},
    c={mark=*,draw=black},
    d={mark=*,draw=black},
    e={mark=*,draw=black},
    f={mark=*,draw=black},
    g={mark=*,draw=black},
    h={mark=*,draw=black},
    i={mark=*,draw=black},
    j={mark=*,draw=black},
    k={mark=*,draw=black},
    l={mark=*,draw=black},
    m={mark=*,draw=black},
    n={mark=*,draw=black},
    o={mark=*,draw=black},
    p={mark=*,draw=black},
    q={mark=*,draw=black},
    r={mark=*,draw=black},
    s={mark=*,draw=black},
    t={mark=*,draw=black}}, xlabel = GBNC-S,
ylabel = CC]
\addplot[draw=blue,pattern=horizontal lines light blue]
 coordinates { (0.13,0.13) (1,1) };

\addplot[scatter,only marks,%
    scatter src=explicit symbolic]%
table[meta=label] {
x       y       label
0.4083	0.5054	w
0.6071	0.6345	b
0.4427	0.4296	c
0.1607	0.1585	d
0.5757	0.5898	e
0.7093	0.6740	f
0.8015	0.8086	g
0.9075	0.9179	h
0.9519	0.9509	i
0.4751	0.7236	j
0.2407	0.2476	k
0.6024	0.6051	l
0.9943	0.9972	m
0.9524	0.9703	n
0.9058	0.9230	o
0.8779	0.9389	p
0.8075	0.8881	q
0.7567	0.7176	r
0.8123	0.8231	s
0.2163	0.2315	t
    };
\end{axis}
\end{tikzpicture} 
&
\begin{tikzpicture}[scale = 0.8]]
\begin{axis}[%
scatter/classes={scatter src = explicit symbolic,%
    w={mark=*,draw=black},
    b={mark=*,draw=black},
    c={mark=*,draw=black},
    d={mark=*,draw=black},
    e={mark=*,draw=black},
    f={mark=*,draw=black},
    g={mark=*,draw=black},
    h={mark=*,draw=black},
    i={mark=*,draw=black},
    j={mark=*,draw=black},
    k={mark=*,draw=black},
    l={mark=*,draw=black},
    m={mark=*,draw=black},
    n={mark=*,draw=black},
    o={mark=*,draw=black},
    p={mark=*,draw=black},
    q={mark=*,draw=black},
    r={mark=*,draw=black},
    s={mark=*,draw=black},
    t={mark=*,draw=black}}, xlabel = BR,
ylabel = CP]
\addplot[draw=blue,pattern=horizontal lines light blue]
 coordinates { (0.1,0.1) (1,1) };

\addplot[scatter,only marks,%
    scatter src=explicit symbolic]%
table[meta=label] {
x       y       label
0.4754	0.5517	w
0.5905	0.9833	b
0.4832	0.6354	c
0.1569	0.8144	d
0.6038	0.9426	e
0.6774	0.9597	f
0.8072	0.8356	g
0.9085	0.9236	h
0.9547	0.9783	i
0.7088	0.9358	j
0.2474	0.2486	k
0.6096	0.6340	l
0.9906	0.9972	m
0.9584	1.0000	n
0.9033	1.0000	o
0.9546	0.9258	p
0.8986	0.9081	q
0.7291	0.8640	r
0.8243	0.9401	s
0.2276	0.2529	t
    };
\end{axis}
\end{tikzpicture}
\\
\\
\\
\begin{tikzpicture}[scale = 0.8]
\begin{axis}[%
scatter/classes={scatter src = explicit symbolic,%
    w={mark=*,draw=black},
    b={mark=*,draw=black},
    c={mark=*,draw=black},
    d={mark=*,draw=black},
    e={mark=*,draw=black},
    f={mark=*,draw=black},
    g={mark=*,draw=black},
    h={mark=*,draw=black},
    i={mark=*,draw=black},
    j={mark=*,draw=black},
    k={mark=*,draw=black},
    l={mark=*,draw=black},
    m={mark=*,draw=black},
    n={mark=*,draw=black},
    o={mark=*,draw=black},
    p={mark=*,draw=black},
    q={mark=*,draw=black},
    r={mark=*,draw=black},
    s={mark=*,draw=black},
    t={mark=*,draw=black}}, xlabel = GBNC-S,
ylabel = CP]
\addplot[draw=blue,pattern=horizontal lines light blue]
 coordinates { (0.15,0.15) (1,1) };

\addplot[scatter,only marks,%
    scatter src=explicit symbolic]%
table[meta=label] {
x       y       label
0.4083	0.5517	w
0.6071	0.9833	b
0.4427	0.6354	c
0.1607	0.8144	d
0.5757	0.9426	e
0.7093	0.9597	f
0.8015	0.8356	g
0.9075	0.9236	h
0.9519	0.9783	i
0.4751	0.9358	j
0.2407	0.2486	k
0.6024	0.6340	l
0.9943	0.9972	m
0.9524	1.0000	n
0.9058	1.0000	o
0.8779	0.9258	p
0.8075	0.9081	q
0.7567	0.8640	r
0.8123	0.9401	s
0.2163	0.2529	t
    };
\end{axis}
\end{tikzpicture}
&
\begin{tikzpicture}[scale = 0.8]
\begin{axis}[%
scatter/classes={scatter src = explicit symbolic,%
    w={mark=*,draw=black},
    b={mark=*,draw=black},
    c={mark=*,draw=black},
    d={mark=*,draw=black},
    e={mark=*,draw=black},
    f={mark=*,draw=black},
    g={mark=*,draw=black},
    h={mark=*,draw=black},
    i={mark=*,draw=black},
    j={mark=*,draw=black},
    k={mark=*,draw=black},
    l={mark=*,draw=black},
    m={mark=*,draw=black},
    n={mark=*,draw=black},
    o={mark=*,draw=black},
    p={mark=*,draw=black},
    q={mark=*,draw=black},
    r={mark=*,draw=black},
    s={mark=*,draw=black},
    t={mark=*,draw=black}}, xlabel = CC,
ylabel = CP]
\addplot[draw=blue,pattern=horizontal lines light blue]
 coordinates { (0.1,0.1) (1,1) };

\addplot[scatter,only marks,%
    scatter src=explicit symbolic]%
table[meta=label] {
x       y       label
0.5054	0.5517	w
0.6345	0.9833	b
0.4296	0.6354	c
0.1585	0.8144	d
0.5898	0.9426	e
0.6740	0.9597	f
0.8086	0.8356	g
0.9179	0.9236	h
0.9509	0.9783	i
0.7236	0.9358	j
0.2476	0.2486	k
0.6051	0.6340	l
0.9972	0.9972	m
0.9703	1.0000	n
0.9230	1.0000	o
0.9389	0.9258	p
0.8881	0.9081	q
0.7176	0.8640	r
0.8231	0.9401	s
0.2315	0.2529	t
    };
\end{axis}
\end{tikzpicture}
\\
\end{tabular}
\caption{Subset 0/1 loss (\textbf{base learner: \textit{Logistic regression}})}
\label{fig:Subset 0/1 loss for pairs with LR}
\end{figure}

\begin{figure}[!ht]
\centering
\begin{tabular}{cc}
\begin{tikzpicture}[scale = 0.8]
\begin{axis}[%
scatter/classes={scatter src = explicit symbolic,%
    w={mark=*,draw=black},
    b={mark=*,draw=black},
    c={mark=*,draw=black},
    d={mark=*,draw=black},
    e={mark=*,draw=black},
    f={mark=*,draw=black},
    g={mark=*,draw=black},
    h={mark=*,draw=black},
    i={mark=*,draw=black},
    j={mark=*,draw=black},
    k={mark=*,draw=black},
    l={mark=*,draw=black},
    m={mark=*,draw=black},
    n={mark=*,draw=black},
    o={mark=*,draw=black},
    p={mark=*,draw=black},
    q={mark=*,draw=black},
    r={mark=*,draw=black},
    s={mark=*,draw=black},
    t={mark=*,draw=black}}, xlabel = GBNC-H,
ylabel = BR]
\addplot[draw=blue,pattern=horizontal lines light blue]
 coordinates { (0.02,0.02) (.8,.8) };

\addplot[scatter,only marks,%
    scatter src=explicit symbolic]%
table[meta=label] {
x       y       label
0.321666667	0.330833333	w
0.430119048	0.453968254	b
0.22598257	0.296232057	c
0.093432673	0.116238477	d
0.342367197	0.382197339	e
0.310443753	0.351827499	f
0.554967283	0.575405254	g
0.529380054	0.60458221	h
0.523719677	0.617654987	i
0.170954829	0.233893791	j
0.225313305	0.345232035	k
0.21193929	0.334365403	l
0.469137466	0.612601078	m
0.294747103	0.332586898	n
0.22710747	0.256543445	o
0.487201907	0.533571401	p
0.339368148	0.332947729	q
0.443807455	0.766925106	r
0.416223095	0.481687082	s
0.035028159	0.240079968	t
    };
\end{axis}
\end{tikzpicture} 
&
\begin{tikzpicture}[scale = 0.8]
\begin{axis}[%
scatter/classes={scatter src = explicit symbolic,%
    w={mark=*,draw=black},
    b={mark=*,draw=black},
    c={mark=*,draw=black},
    d={mark=*,draw=black},
    e={mark=*,draw=black},
    f={mark=*,draw=black},
    g={mark=*,draw=black},
    h={mark=*,draw=black},
    i={mark=*,draw=black},
    j={mark=*,draw=black},
    k={mark=*,draw=black},
    l={mark=*,draw=black},
    m={mark=*,draw=black},
    n={mark=*,draw=black},
    o={mark=*,draw=black},
    p={mark=*,draw=black},
    q={mark=*,draw=black},
    r={mark=*,draw=black},
    s={mark=*,draw=black},
    t={mark=*,draw=black}}, xlabel = BR,
ylabel = CC]
\addplot[draw=blue,pattern=horizontal lines light blue]
 coordinates { (0.1,0.1) (.8,.8) };

\addplot[scatter,only marks,%
    scatter src=explicit symbolic]%
table[meta=label] {
x       y       label
0.330833333	0.341875	w
0.453968254	0.434484127	b
0.296232057	0.294933356	c
0.116238477	0.124210435	d
0.382197339	0.380071405	e
0.351827499	0.351483102	f
0.575405254	0.578180751	g
0.60458221	0.627493261	h
0.617654987	0.619002695	i
0.233893791	0.2351037	j
0.345232035	0.397136178	k
0.334365403	0.368945935	l
0.612601078	0.644541779	m
0.332586898	0.332948975	n
0.256543445	0.258708079	o
0.533571401	0.571951403	p
0.332947729	0.34099991	q
0.766925106	0.325981519	r
0.481687082	0.480410271	s
0.240079968	0.200284541	t
    };
\end{axis}
\end{tikzpicture}
\\
\\
\\
\begin{tikzpicture}[scale = 0.8]
\begin{axis}[%
scatter/classes={scatter src = explicit symbolic,%
    w={mark=*,draw=black},
    b={mark=*,draw=black},
    c={mark=*,draw=black},
    d={mark=*,draw=black},
    e={mark=*,draw=black},
    f={mark=*,draw=black},
    g={mark=*,draw=black},
    h={mark=*,draw=black},
    i={mark=*,draw=black},
    j={mark=*,draw=black},
    k={mark=*,draw=black},
    l={mark=*,draw=black},
    m={mark=*,draw=black},
    n={mark=*,draw=black},
    o={mark=*,draw=black},
    p={mark=*,draw=black},
    q={mark=*,draw=black},
    r={mark=*,draw=black},
    s={mark=*,draw=black},
    t={mark=*,draw=black}}, xlabel = GBNC-H,
ylabel = CC]
\addplot[draw=blue,pattern=horizontal lines light blue]
 coordinates { (0.02,0.02) (.65,.65) };

\addplot[scatter,only marks,%
    scatter src=explicit symbolic]%
table[meta=label] {
x       y       label
0.321666667	0.341875	w
0.430119048	0.434484127	b
0.22598257	0.294933356	c
0.093432673	0.124210435	d
0.342367197	0.380071405	e
0.310443753	0.351483102	f
0.554967283	0.578180751	g
0.529380054	0.627493261	h
0.523719677	0.619002695	i
0.170954829	0.2351037	j
0.225313305	0.397136178	k
0.21193929	0.368945935	l
0.469137466	0.644541779	m
0.294747103	0.332948975	n
0.22710747	0.258708079	o
0.487201907	0.571951403	p
0.339368148	0.34099991	q
0.443807455	0.325981519	r
0.416223095	0.480410271	s
0.035028159	0.200284541	t
    };
\end{axis}
\end{tikzpicture} 
&
\begin{tikzpicture}[scale = 0.8]
\begin{axis}[%
scatter/classes={scatter src = explicit symbolic,%
    w={mark=*,draw=black},
    b={mark=*,draw=black},
    c={mark=*,draw=black},
    d={mark=*,draw=black},
    e={mark=*,draw=black},
    f={mark=*,draw=black},
    g={mark=*,draw=black},
    h={mark=*,draw=black},
    i={mark=*,draw=black},
    j={mark=*,draw=black},
    k={mark=*,draw=black},
    l={mark=*,draw=black},
    m={mark=*,draw=black},
    n={mark=*,draw=black},
    o={mark=*,draw=black},
    p={mark=*,draw=black},
    q={mark=*,draw=black},
    r={mark=*,draw=black},
    s={mark=*,draw=black},
    t={mark=*,draw=black}}, xlabel = BR,
ylabel = CP]
\addplot[draw=blue,pattern=horizontal lines light blue]
 coordinates { (0.1,0.1) (.8,.8) };

\addplot[scatter,only marks,%
    scatter src=explicit symbolic]%
table[meta=label] {
x           y           label
0.330833333	0.27875	    w
0.453968254	0.690634921	b
0.296232057	0.31254272	c
0.116238477	0.455363139	d
0.382197339	0.569133398	e
0.351827499	0.518297785	f
0.575405254	0.591060944	g
0.60458221	0.497169811	h
0.617654987	0.484770889	i
0.233893791	0.376948496	j
0.345232035	0.18610062	k
0.334365403	0.200948406	l
0.612601078	0.406469003	m
0.332586898	0.700729724	n
0.256543445	0.573193598	o
0.533571401	0.59207667	p
0.332947729	0.575226601	q
0.766925106	0.6062903	r
0.481687082	0.633161969	s
0.240079968	0.078188256	t
    };
\end{axis}
\end{tikzpicture}
\\
\\
\\
\begin{tikzpicture}[scale = 0.8]
\begin{axis}[%
scatter/classes={scatter src = explicit symbolic,%
    w={mark=*,draw=black},
    b={mark=*,draw=black},
    c={mark=*,draw=black},
    d={mark=*,draw=black},
    e={mark=*,draw=black},
    f={mark=*,draw=black},
    g={mark=*,draw=black},
    h={mark=*,draw=black},
    i={mark=*,draw=black},
    j={mark=*,draw=black},
    k={mark=*,draw=black},
    l={mark=*,draw=black},
    m={mark=*,draw=black},
    n={mark=*,draw=black},
    o={mark=*,draw=black},
    p={mark=*,draw=black},
    q={mark=*,draw=black},
    r={mark=*,draw=black},
    s={mark=*,draw=black},
    t={mark=*,draw=black}}, xlabel = GBNC-H,
ylabel = CP]
\addplot[draw=blue,pattern=horizontal lines light blue]
 coordinates { (0.02,0.02) (.65,.65) };

\addplot[scatter,only marks,%
    scatter src=explicit symbolic]%
table[meta=label] {
x           y           label
0.321666667	0.27875	    w
0.430119048	0.690634921	b
0.22598257	0.31254272	c
0.093432673	0.455363139	d
0.342367197	0.569133398	e
0.310443753	0.518297785	f
0.554967283	0.591060944	g
0.529380054	0.497169811	h
0.523719677	0.484770889	i
0.170954829	0.376948496	j
0.225313305	0.18610062	k
0.21193929	0.200948406	l
0.469137466	0.406469003	m
0.294747103	0.700729724	n
0.22710747	0.573193598	o
0.487201907	0.59207667	p
0.339368148	0.575226601	q
0.443807455	0.6062903	r
0.416223095	0.633161969	s
0.035028159	0.078188256	t
    };
\end{axis}
\end{tikzpicture}
&
\begin{tikzpicture}[scale = 0.8]
\begin{axis}[%
scatter/classes={scatter src = explicit symbolic,%
    w={mark=*,draw=black},
    b={mark=*,draw=black},
    c={mark=*,draw=black},
    d={mark=*,draw=black},
    e={mark=*,draw=black},
    f={mark=*,draw=black},
    g={mark=*,draw=black},
    h={mark=*,draw=black},
    i={mark=*,draw=black},
    j={mark=*,draw=black},
    k={mark=*,draw=black},
    l={mark=*,draw=black},
    m={mark=*,draw=black},
    n={mark=*,draw=black},
    o={mark=*,draw=black},
    p={mark=*,draw=black},
    q={mark=*,draw=black},
    r={mark=*,draw=black},
    s={mark=*,draw=black},
    t={mark=*,draw=black}}, xlabel = CC,
ylabel = CP]
\addplot[draw=blue,pattern=horizontal lines light blue]
 coordinates { (0.15,0.15) (.81,.81) };

\addplot[scatter,only marks,%
    scatter src=explicit symbolic]%
table[meta=label] {
x           y           label
0.341875	0.27875	    w
0.434484127	0.690634921	b
0.294933356	0.31254272	c
0.124210435	0.455363139	d
0.380071405	0.569133398	e
0.351483102	0.518297785	f
0.578180751	0.591060944	g
0.627493261	0.497169811	h
0.619002695	0.484770889	i
0.2351037	0.376948496	j
0.397136178	0.18610062	k
0.368945935	0.200948406	l
0.644541779	0.406469003	m
0.332948975	0.700729724	n
0.258708079	0.573193598	o
0.571951403	0.59207667	p
0.34099991	0.575226601	q
0.325981519	0.6062903	r
0.480410271	0.633161969	s
0.200284541	0.078188256	t
    };
\end{axis}
\end{tikzpicture}
\end{tabular}
\caption{Hamming loss (\textbf{base learner: \textit{Naive Bayes}})}
\label{fig:Hamming loss for pairs with NB}
\end{figure}
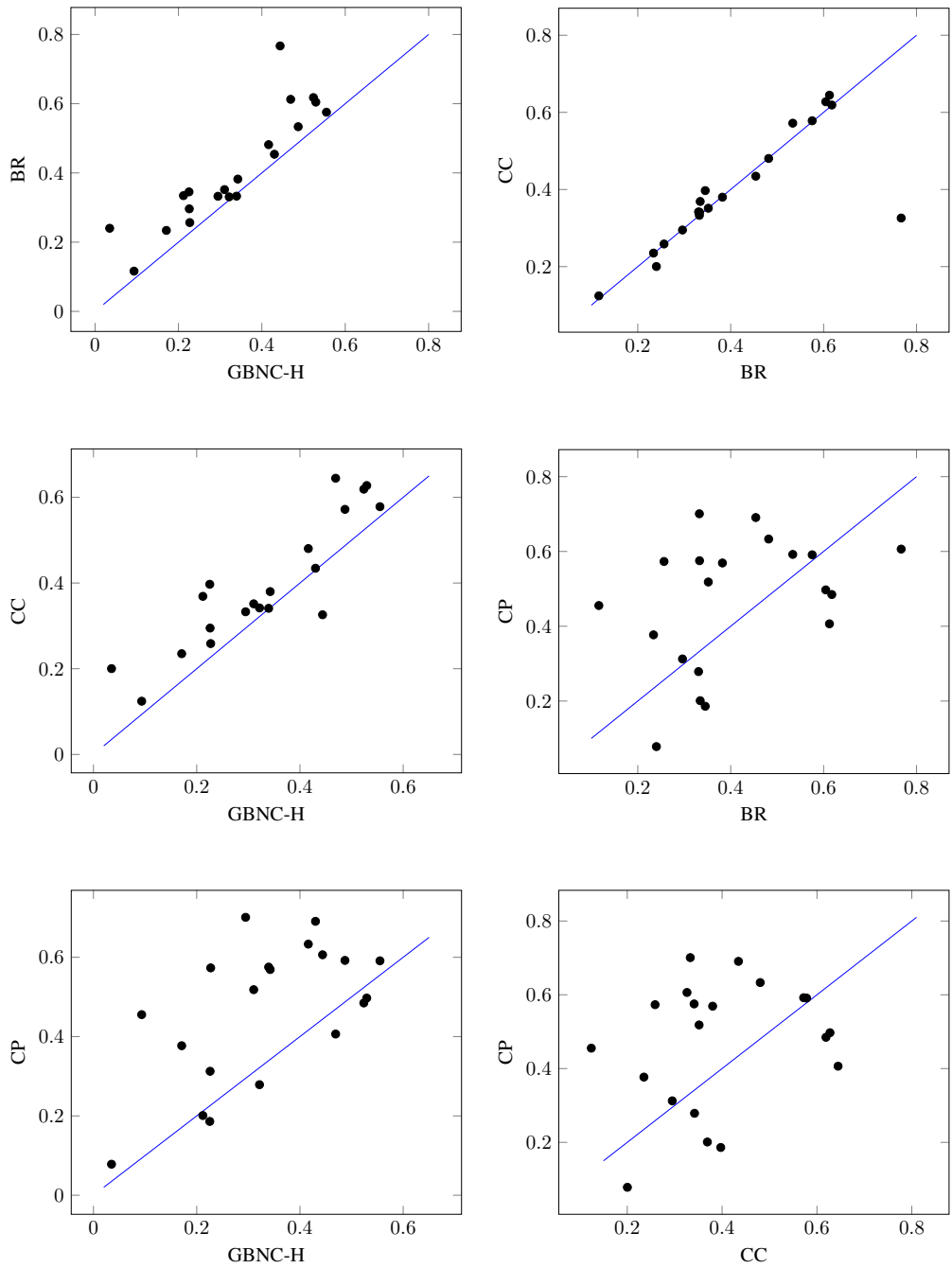

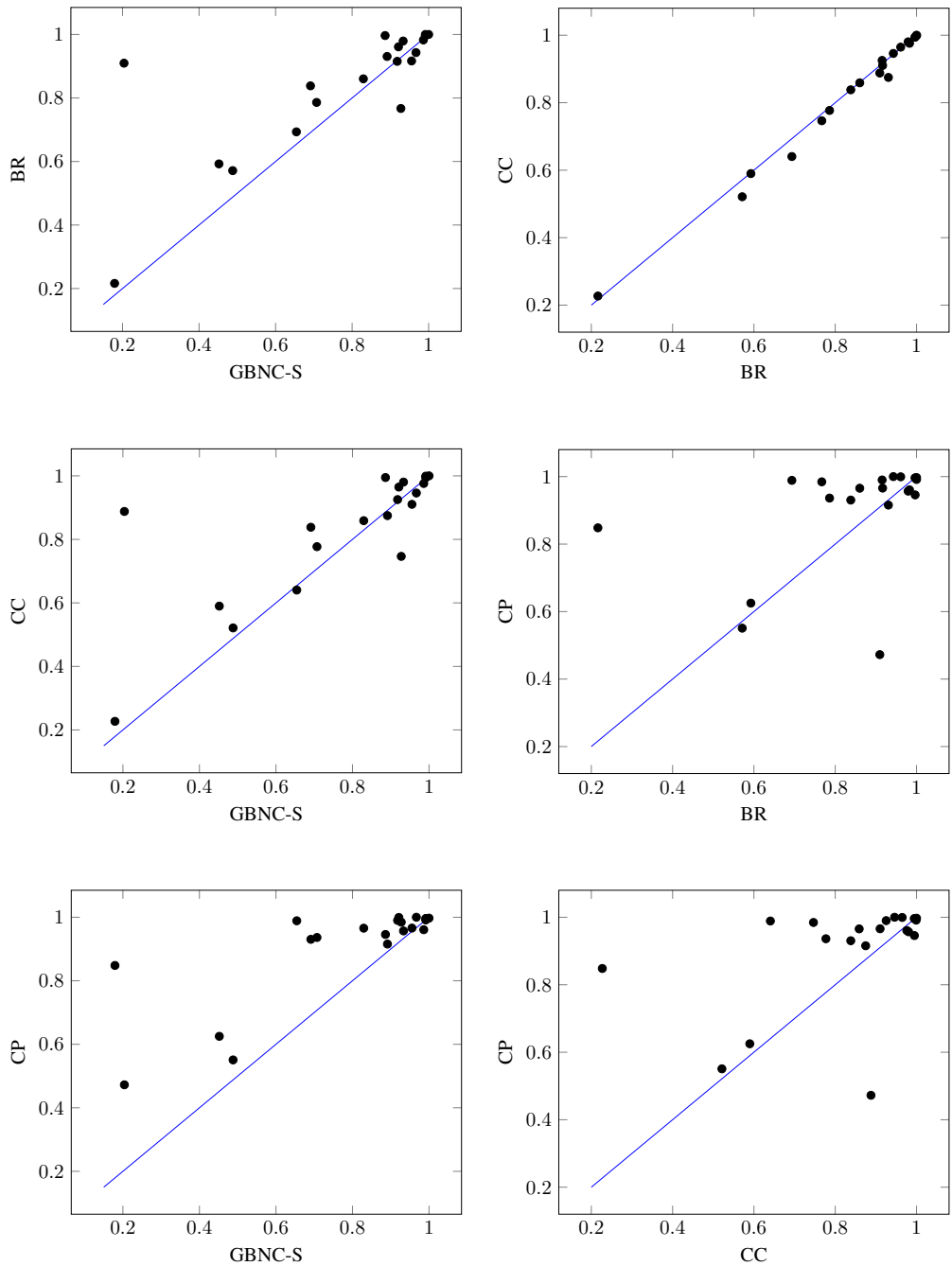
\begin{figure}[!ht]
\centering
\begin{tabular}{cc}
\begin{tikzpicture}[scale = 0.8]
\begin{axis}[%
scatter/classes={scatter src = explicit symbolic,%
    w={mark=*,draw=black},
    b={mark=*,draw=black},
    c={mark=*,draw=black},
    d={mark=*,draw=black},
    e={mark=*,draw=black},
    f={mark=*,draw=black},
    g={mark=*,draw=black},
    h={mark=*,draw=black},
    i={mark=*,draw=black},
    j={mark=*,draw=black},
    k={mark=*,draw=black},
    l={mark=*,draw=black},
    m={mark=*,draw=black},
    n={mark=*,draw=black},
    o={mark=*,draw=black},
    p={mark=*,draw=black},
    q={mark=*,draw=black},
    r={mark=*,draw=black},
    s={mark=*,draw=black},
    t={mark=*,draw=black}}, xlabel = GBNC-S,
ylabel = BR]
\addplot[draw=blue,pattern=horizontal lines light blue]
 coordinates { (0.15,0.15) (1,1) };

\addplot[scatter,only marks,%
    scatter src=explicit symbolic]%
table[meta=label] {
x           y           label
0.487916667	0.57125	    w
0.654365079	0.693333333	b
0.45196514	0.592464115	c
0.179210842	0.216211514	d
0.707189224	0.786043492	e
0.829399198	0.860303931	f
0.933390148	0.97936262	g
0.991509434	1	        h
0.991509434	0.996226415	i
0.691218528	0.838213581	j
0.891411435	0.930757875	k
0.886218896	0.996563777	l
1	        1	        m
0.966934046	0.942959002	n
0.917926829	0.915365854	o
0.9862824	0.982711842	p
0.955422934	0.916556825	q
0.927466278	0.766925106	r
0.921122605	0.961047238	s
0.20387714	0.909617774	t
    };
\end{axis}
\end{tikzpicture} 
&
\begin{tikzpicture}[scale = 0.8]
\begin{axis}[%
scatter/classes={scatter src = explicit symbolic,%
    w={mark=*,draw=black},
    b={mark=*,draw=black},
    c={mark=*,draw=black},
    d={mark=*,draw=black},
    e={mark=*,draw=black},
    f={mark=*,draw=black},
    g={mark=*,draw=black},
    h={mark=*,draw=black},
    i={mark=*,draw=black},
    j={mark=*,draw=black},
    k={mark=*,draw=black},
    l={mark=*,draw=black},
    m={mark=*,draw=black},
    n={mark=*,draw=black},
    o={mark=*,draw=black},
    p={mark=*,draw=black},
    q={mark=*,draw=black},
    r={mark=*,draw=black},
    s={mark=*,draw=black},
    t={mark=*,draw=black}}, xlabel = BR,
ylabel = CC]
\addplot[draw=blue,pattern=horizontal lines light blue]
 coordinates { (0.2,0.2) (1,1) };

\addplot[scatter,only marks,%
    scatter src=explicit symbolic]%
table[meta=label] {
x           y           label
0.57125	    0.52125	    w
0.693333333	0.64047619	b
0.592464115	0.589866712	c
0.216211514	0.227054801	d
0.786043492	0.777052905	e
0.860303931	0.858910018	f
0.97936262	0.980430084	g
1	        0.999056604	h
0.996226415	0.994339623	i
0.838213581	0.838212218	j
0.930757875	0.874866572	k
0.996563777	0.994731017	l
1	        1	        m
0.942959002	0.945989305	n
0.915365854	0.925304878	o
0.982711842	0.976020017	p
0.916556825	0.910333583	q
0.766925106	0.746456807	r
0.961047238	0.964660847	s
0.909617774	0.887922158	t
    };
\end{axis}
\end{tikzpicture}
\\
\\
\\
\begin{tikzpicture}[scale = 0.8]
\begin{axis}[%
scatter/classes={scatter src = explicit symbolic,%
    w={mark=*,draw=black},
    b={mark=*,draw=black},
    c={mark=*,draw=black},
    d={mark=*,draw=black},
    e={mark=*,draw=black},
    f={mark=*,draw=black},
    g={mark=*,draw=black},
    h={mark=*,draw=black},
    i={mark=*,draw=black},
    j={mark=*,draw=black},
    k={mark=*,draw=black},
    l={mark=*,draw=black},
    m={mark=*,draw=black},
    n={mark=*,draw=black},
    o={mark=*,draw=black},
    p={mark=*,draw=black},
    q={mark=*,draw=black},
    r={mark=*,draw=black},
    s={mark=*,draw=black},
    t={mark=*,draw=black}}, xlabel = GBNC-S,
ylabel = CC]
\addplot[draw=blue,pattern=horizontal lines light blue]
 coordinates { (0.15,0.15) (1,1) };

\addplot[scatter,only marks,%
    scatter src=explicit symbolic]%
table[meta=label] {
x           y           label
0.487916667	0.52125	    w
0.654365079	0.64047619	b
0.45196514	0.589866712	c
0.179210842	0.227054801	d
0.707189224	0.777052905	e
0.829399198	0.858910018	f
0.933390148	0.980430084	g
0.991509434	0.999056604	h
0.991509434	0.994339623	i
0.691218528	0.838212218	j
0.891411435	0.874866572	k
0.886218896	0.994731017	l
1	        1	        m
0.966934046	0.945989305	n
0.917926829	0.925304878	o
0.9862824	0.976020017	p
0.955422934	0.910333583	q
0.927466278	0.746456807	r
0.921122605	0.964660847	s
0.20387714	0.887922158	t
    };
\end{axis}
\end{tikzpicture} 
&
\begin{tikzpicture}[scale = 0.8]
\begin{axis}[%
scatter/classes={scatter src = explicit symbolic,%
    w={mark=*,draw=black},
    b={mark=*,draw=black},
    c={mark=*,draw=black},
    d={mark=*,draw=black},
    e={mark=*,draw=black},
    f={mark=*,draw=black},
    g={mark=*,draw=black},
    h={mark=*,draw=black},
    i={mark=*,draw=black},
    j={mark=*,draw=black},
    k={mark=*,draw=black},
    l={mark=*,draw=black},
    m={mark=*,draw=black},
    n={mark=*,draw=black},
    o={mark=*,draw=black},
    p={mark=*,draw=black},
    q={mark=*,draw=black},
    r={mark=*,draw=black},
    s={mark=*,draw=black},
    t={mark=*,draw=black}}, xlabel = BR,
ylabel = CP]
\addplot[draw=blue,pattern=horizontal lines light blue]
 coordinates { (0.2,0.2) (1,1) };

\addplot[scatter,only marks,%
    scatter src=explicit symbolic]%
table[meta=label] {
x           y           label
0.57125   	0.550833333	w
0.693333333	0.988888889	b
0.592464115	0.625085441	c
0.216211514	0.848214322	d
0.786043492	0.936335605	e
0.860303931	0.965567786	f
0.97936262	0.956944766	g
1	        0.991509434	h
0.996226415	0.996226415	i
0.838213581	0.93045527	j
0.930757875	0.915554468	k
0.996563777	0.945781466	l
1	        0.997169811	m
0.942959002	1	        n
0.915365854	0.990060976	o
0.982711842	0.96074141	p
0.916556825	0.965928665	q
0.766925106	0.984581268	r
0.961047238	0.9994093	s
0.909617774	0.472416091	t
    };
\end{axis}
\end{tikzpicture}
\\
\\
\\
\begin{tikzpicture}[scale = 0.8]
\begin{axis}[%
scatter/classes={scatter src = explicit symbolic,%
    w={mark=*,draw=black},
    b={mark=*,draw=black},
    c={mark=*,draw=black},
    d={mark=*,draw=black},
    e={mark=*,draw=black},
    f={mark=*,draw=black},
    g={mark=*,draw=black},
    h={mark=*,draw=black},
    i={mark=*,draw=black},
    j={mark=*,draw=black},
    k={mark=*,draw=black},
    l={mark=*,draw=black},
    m={mark=*,draw=black},
    n={mark=*,draw=black},
    o={mark=*,draw=black},
    p={mark=*,draw=black},
    q={mark=*,draw=black},
    r={mark=*,draw=black},
    s={mark=*,draw=black},
    t={mark=*,draw=black}}, xlabel = GBNC-S,
ylabel = CP]
\addplot[draw=blue,pattern=horizontal lines light blue]
 coordinates { (0.15,0.15) (1,1) };

\addplot[scatter,only marks,%
    scatter src=explicit symbolic]%
table[meta=label] {
x           y           label
0.487916667	0.550833333	w
0.654365079	0.988888889	b
0.45196514	0.625085441	c
0.179210842	0.848214322	d
0.707189224	0.936335605	e
0.829399198	0.965567786	f
0.933390148	0.956944766	g
0.991509434	0.991509434	h
0.991509434	0.996226415	i
0.691218528	0.93045527	j
0.891411435	0.915554468	k
0.886218896	0.945781466	l
1	        0.997169811	m
0.966934046	1	        n
0.917926829	0.990060976	o
0.9862824	0.96074141	p
0.955422934	0.965928665	q
0.927466278	0.984581268	r
0.921122605	0.9994093	s
0.20387714	0.472416091	t
    };
\end{axis}
\end{tikzpicture}
&
\begin{tikzpicture}[scale = 0.8]
\begin{axis}[%
scatter/classes={scatter src = explicit symbolic,%
    w={mark=*,draw=black},
    b={mark=*,draw=black},
    c={mark=*,draw=black},
    d={mark=*,draw=black},
    e={mark=*,draw=black},
    f={mark=*,draw=black},
    g={mark=*,draw=black},
    h={mark=*,draw=black},
    i={mark=*,draw=black},
    j={mark=*,draw=black},
    k={mark=*,draw=black},
    l={mark=*,draw=black},
    m={mark=*,draw=black},
    n={mark=*,draw=black},
    o={mark=*,draw=black},
    p={mark=*,draw=black},
    q={mark=*,draw=black},
    r={mark=*,draw=black},
    s={mark=*,draw=black},
    t={mark=*,draw=black}}, xlabel = CC,
ylabel = CP]
\addplot[draw=blue,pattern=horizontal lines light blue]
 coordinates { (0.2,0.2) (1,1) };

\addplot[scatter,only marks,%
    scatter src=explicit symbolic]%
table[meta=label] {
x           y           label
0.52125	    0.550833333	w
0.64047619	0.988888889	b
0.589866712	0.625085441	c
0.227054801	0.848214322	d
0.777052905	0.936335605	e
0.858910018	0.965567786	f
0.980430084	0.956944766	g
0.999056604	0.991509434	h
0.994339623	0.996226415	i
0.838212218	0.93045527	j
0.874866572	0.915554468	k
0.994731017	0.945781466	l
1	        0.997169811	m
0.945989305	1        	n
0.925304878	0.990060976	o
0.976020017	0.96074141	p
0.910333583	0.965928665	q
0.746456807	0.984581268	r
0.964660847	0.9994093	s
0.887922158	0.472416091	t
    };
\end{axis}
\end{tikzpicture}
\end{tabular}
\caption{Subset 0/1 loss (\textbf{base learner: \textit{Naive Bayes}})}
\label{fig:Subset 0/1 loss for pairs with NB}
\end{figure}

Hamming losses and their ranks provided by the classifiers are given in table \ref{tab:lr_hs} and \ref{tab:nb_hs}. Subset 0/1 losses and their ranks provided by the classifiers are given in table \ref{tab:lr_ss} and \ref{tab:nb_ss}. Scatter plots for the losses provided by pairs of classifiers are given in Figure \ref{fig:Hamming loss for pairs with LR}--\ref{fig:Subset 0/1 loss for pairs with NB}. Each black point illustrates losses provided by the classifiers labeled on the horizontal axis and the vertical axis on one data set. The differences provided by pairs of classifiers are illustrated by the horizontal distances between (black) points and the blue line $y =x$. Points lie on the left side of $y =x$ indicate that classifiers labeled on the horizontal axis are better than ones labeled on the vertical axis, and points lie on the right side of $y =x$ indicate that classifiers labeled on the vertical axis are better than ones labeled in the horizontal axis. Points lie far away from  $y =x$ suggest visible differences. 

\begin{table}[!ht]
    \centering
    \caption{Hamming loss (mean $\pm$ std.) of each MDC approach (\textbf{base learner: \textit{logisitic regression}}).}
    \begin{tabular}{p{5em}|p{7.15em}p{7.45em}p{7.65em}p{7.15em}}
    \toprule
    \multirow{2}{*}{Data Set} & \multicolumn{4}{c}{Hamming loss (in \%)} \\ \cmidrule(r){2-5}
    & \multicolumn{1}{c}{GBNC-H} & BR & CC & CP  \\ \hline
Edm	& \bfseries	26.54	$	\pm	$	9.57	(1.0)	&	27.65	$	\pm	$	7.95	(3.0)	&	27.23	$	\pm	$	6.95	(2.0)	&	28.23	$	\pm	$	8.10	(4.0) \\
Jura	&	37.32	$	\pm	$	4.48	(2.0)	&	\bfseries 35.51	$	\pm	$	5.06	(1.0)	&	39.81	$	\pm	$	8.10	(3.0)	&	67.84	$	\pm	$	7.23	(4.0) \\
Enb	&	\bfseries 22.20	$	\pm	$	4.59	(1.5)	&	24.16	$	\pm	$	4.44	(3.0)	&	\bfseries 22.20	$	\pm	$	5.29	(1.5)	&	31.77	$	\pm	$	1.89	(4.0) \\
Voice	&	8.21	$	\pm	$	1.18	(3.0)	&	\bfseries 8.08	$	\pm	$	\bfseries 1.05	(1.5)	&	8.08	$	\pm	$	0.67	(1.5)	&	41.69	$	\pm	$	1.30	(4.0) \\
Song	&\bfseries 	24.33	$	\pm	$	2.35	(1.0)	&	25.74	$	\pm	$	2.86	(2.0)	&	26.46	$	\pm	$	5.49	(3.0)	&	49.30	$	\pm	$	3.32	(4.0) \\
Adult	&	32.46	$	\pm	$	0.77	(3.0)	& \bfseries	28.26	$	\pm	$	0.47	(1.0)	&	28.46	$	\pm	$	0.42	(2.0)	&	41.40	$	\pm	$	0.61	(4.0) \\
Default	&	\bfseries 33.29	$	\pm	$	0.42	(1.0)	&	33.48	$	\pm	$	0.59	(3.0)	&	33.39	$	\pm	$	0.42	(2.0)	&	43.94	$	\pm	$	0.31	(4.0) \\

Flickr	&	21.74	$	\pm	$	0.57	(3.0)	&	\bfseries 20.22	$	\pm	$	0.69	(1.0)	&	20.46	$	\pm	$	0.47	(2.0)	&	49.21	$	\pm	$	0.72	(4.0) \\
Fera	&	\bfseries 37.76	$	\pm	$	0.64	(1.0)	&	38.82	$	\pm	$	0.96	(2.0)	&	39.23	$	\pm	$	0.70	(3.00)	&	52.97	$	\pm	$	2.41	(4.0) \\

\hline

WQplants	& \bfseries	34.61	$	\pm	$	1.67	(1.0)	&	34.65	$	\pm	$	2.18	(2.0)	&	35.42	$	\pm	$	2.02	(3.0)	&	38.06	$	\pm	$	3.05	(4.0) \\
WQanimals	&	\bfseries 36.85	$	\pm	$	1.35	(1.0)	&	36.98	$	\pm	$	1.97	(2.0)	&	38.23	$	\pm	$	0.87	(3.0)	&	43.07	$	\pm	$	2.68	(4.0) \\

Thyroid	&	\bfseries 3.38	$	\pm	$	0.14	(1.0)	&	3.52	$	\pm	$	0.19	(3.0)	&	3.44	$	\pm	$	0.17	(2.0)	&	3.89	$	\pm	$	0.16	(4.0) \\

\hline
Rf1	&	\bfseries 9.53	$	\pm	$	0.65	(1.0)	&	16.10	$	\pm	$	0.67	(2.0)	&	16.33	$	\pm	$	0.42	(3.0)	&	36.49	$	\pm	$	0.79	(4.0) \\
Pain	&	\bfseries 4.70	$	\pm	$	0.33	(1.0)	&	4.74	$	\pm	$	0.32	(2.0)	&	4.91	$	\pm	$	0.37	(3.0)	&	5.24	$	\pm	$	0.46	(4.0) \\
Disfa	&	\bfseries 10.30	$	\pm	$	0.36	(1.0)	&	10.58	$	\pm	$	0.37	(2.0)	&	10.63	$	\pm	$	0.30	(3.0)	&	13.08	$	\pm	$	0.60	(4.0) \\
WaterQuality	&	\bfseries 35.53	$	\pm	$	1.24	(1.0)	&	36.10	$	\pm	$	1.12	(2.0)	&	36.50	$	\pm	$	0.95	(3.0)	&	40.92	$	\pm	$	1.58	(4.0) \\
Oes97	&	\bfseries 27.61	$	\pm	$	1.41	(1.0)	&	28.24	$	\pm	$	1.68	(2.0)	&	29.35	$	\pm	$	1.78	(3.0)	&	45.59	$	\pm	$	3.36	(4.0) \\
Oes10	&	19.55	$	\pm	$	1.80	(2.0)	& \bfseries	19.21	$	\pm	$	1.98	(1.0)	&	20.80	$	\pm	$	1.73	(3.0)	&	38.40	$	\pm	$	3.30	(4.0) \\
Scm20d	&	\bfseries 31.45	$	\pm	$	0.84	(1.0)	&	36.04	$	\pm	$	0.71	(2.0)	&	38.15	$	\pm	$	0.96	(3.0)	&	57.87	$	\pm	$	0.65	(4.0) \\
Scm1d	&	\bfseries 18.07	$	\pm	$	0.38	(1.0)	&	23.42	$	\pm	$	0.79	(2.0)	&	25.63	$	\pm	$	0.95	(3.0)	&	55.59	$	\pm	$	0.80	(4.0) \\
\hline 
Ave. rank	&				\bfseries			1.43		&							1.98		&							2.60 		&						4.00	\\
    \bottomrule
    \end{tabular}
    \label{tab:lr_hs}
\end{table}

\begin{table}[!ht]
    \centering
    \caption{Hamming loss (mean $\pm$ std.) of each MDC approach (\textbf{base learner: \textit{Naive Bayes}}).}
    \label{tab:nb_hs}
    \begin{tabular}{p{5em}|p{7.15em}p{7.45em}p{7.65em}p{7.15em}}
    \toprule
    \multirow{2}{*}{Data Set} & \multicolumn{4}{c}{Hamming loss (in \%)} \\ \cmidrule(r){2-5}
    & \multicolumn{1}{c}{GBNC-H} & BR & CC & CP  \\ \hline
Edm	&	32.17	$	\pm	$	6.55	(2.0)	&	33.08	$	\pm	$	3.67	(3.0)	&	34.19	$	\pm	$	7.64	(4.0)	&	\bfseries 27.88	$	\pm	$	7.41	(1.0) \\
Jura	&	\bfseries 43.01	$	\pm	$	6.83	(1.0)	&	45.40	$	\pm	$	4.45	(3.0)	&	43.45	$	\pm	$	4.45	(2.0)	&	69.06	$	\pm	$	4.25	(4.0) \\
Enb	&	\bfseries 22.60	$	\pm	$	2.21	(1.0)	&	29.62	$	\pm	$	2.81	(3.0)	&	29.49	$	\pm	$	3.18	(2.0)	&	31.25	$	\pm	$	2.91	(4.0) \\
Voice	&	\bfseries 9.34	$	\pm	$	0.85	(1.0)	&	11.62	$	\pm	$	1.50	(2.0)	&	12.42	$	\pm	$	1.75	(3.0)	&	45.54	$	\pm	$	1.52	(4.0) \\
Song	& \bfseries	34.24	$	\pm	$	3.17	(1.0)	&	38.22	$	\pm	$	3.14	(3.0)	&	38.01	$	\pm	$	2.36	(2.0)	&	56.91	$	\pm	$	5.67	(4.0) \\
Adult	&	44.38	$	\pm	$	1.03	(2.0)	&	76.69	$	\pm	$	3.14	(4.0)	& \bfseries	32.60	$	\pm	$	0.79	(1.0)	&	60.63	$	\pm	$	1.19	(3.0) \\
Default	&	\bfseries 41.62	$	\pm	$	2.01	(1.0)	&	48.17	$	\pm	$	2.74	(3.0)	&	48.04	$	\pm	$	2.49	(2.0)	&	63.32	$	\pm	$	1.08	(4.0) \\
Flickr	& \bfseries	31.04	$	\pm	$	0.65	(1.0)	&	35.18	$	\pm	$	0.69	(3.0)	&	35.15	$	\pm	$	0.73	(2.0)	&	51.83	$	\pm	$	0.56	(4.0) \\
Fera	& \bfseries	55.50	$	\pm	$	0.76	(1.0)	&	57.54	$	\pm	$	0.38	(2.0)	&	57.82	$	\pm	$	0.38	(3.0)	&	59.11	$	\pm	$	0.55	(4.0) \\

\hline

WQplants	&	52.94	$	\pm	$	4.31	(2.0)	&	60.46	$	\pm	$	3.35	(3.0)	&	62.75	$	\pm	$	3.86	(4.0)	&	\bfseries 49.72	$	\pm	$	2.39	(1.0) \\
WQanimals	&	52.37	$	\pm	$	1.43	(2.0)	&	61.77	$	\pm	$	0.89	(3.0)	&	61.90	$	\pm	$	1.04	(4.0)	&	\bfseries 48.48	$	\pm	$	1.79	(1.0) \\
Thyroid	&	\bfseries 3.50	$	\pm	$	0.35	(1.0)	&	24.01	$	\pm	$	1.68	(4.0)	&	20.03	$	\pm	$	0.68	(3.0)	&	7.82	$	\pm	$	0.34	(2.0) \\

\hline

Rf1	&	\bfseries 17.10	$	\pm	$	0.60	(1.0)	&	23.39	$	\pm	$	0.40	(2.0)	&	23.51	$	\pm	$	0.39	(3.0)	&	37.69	$	\pm	$	0.90	(4.0) \\
Pain	&	22.53	$	\pm	$	0.49	(2.0)	&	34.52	$	\pm	$	1.03	(3.0)	&	39.71	$	\pm	$	2.07	(4.0)	&	\bfseries 18.61	$	\pm	$	0.54	(1.0) \\
Disfa	&	21.19	$	\pm	$	1.03	(2.0)	&	33.44	$	\pm	$	0.57	(3.0)	&	36.89	$	\pm	$	0.68	(4.0)	&	\bfseries 20.09	$	\pm	$	0.22	(1.0) \\
WaterQuality	&	46.91	$	\pm	$	1.88	(2.0)	&	61.26	$	\pm	$	1.20	(3.0)	&	64.45	$	\pm	$	1.25	(4.0)	&	\bfseries 40.65	$	\pm	$	1.57	(1.0)\\
Oes97	& \bfseries	29.47	$	\pm	$	1.95	(1.0)	&	33.26	$	\pm	$	1.60	(2.0)	&	33.29	$	\pm	$	1.62	(3.0)	&	70.07	$	\pm	$	4.16	(4.0)\\
Oes10	& \bfseries	22.71	$	\pm	$	1.55	(1.0)	&	25.65	$	\pm	$	2.15	(2.0)	&	25.87	$	\pm	$	2.08	(3.0)	&	57.32	$	\pm	$	5.16	(4.0) \\
Scm20d	& \bfseries	48.72	$	\pm	$	0.75	(1.0)	&	53.36	$	\pm	$	0.68	(2.0)	&	57.20	$	\pm	$	0.76	(3.0)	&	59.21	$	\pm	$	0.98	(4.0)\\
Scm1d	&	33.94	$	\pm	$	0.67	(2.0)	& \bfseries	33.29	$	\pm	$	0.64	(1.0)	&	34.10	$	\pm	$	0.68	(3.0)	&	57.52	$	\pm	$	1.13	(4.0) \\
\hline
Ave. rank	&			\bfseries				1.40		&							2.70		&							2.95		&							2.95	\\
    \bottomrule
    \end{tabular}
\end{table}

\begin{table}
    \centering
    \caption{Subset 0/1 loss (mean $\pm$ std.) of each MDC approach (\textbf{base learner: \textit{logisitic regression}}).}
    \begin{tabular}{p{5em}|p{7.15em}p{7.45em}p{7.65em}p{7.15em}}
    \toprule
    \multirow{2}{*}{Data Set} & \multicolumn{4}{c}{Subset 0/1 loss (in \%)} \\ \cmidrule(r){2-5}
    & \multicolumn{1}{c}{GBNC-S} & BR & CC & CP  \\ \hline
Edm	& \small 	\bfseries 40.83	$	\pm	$	11.73	(1.0)	&	 \small 47.54	$	\pm	$	12.49	(2.0)	&	 \small 50.54	$	\pm	$	15.58	(3.0)	&	 \small  55.17	$	\pm	$	14.99	(4.0) \\
Jura	&	60.71	$	\pm	$	5.75	(2.0)	& \bfseries	59.05	$	\pm	$	5.86	(1.0)	&	63.45	$	\pm	$	8.29	(3.0)	&	98.33	$	\pm	$	4.16	(4.0) \\
Enb	&	44.27	$	\pm	$	8.86	(2.0)	&	48.32	$	\pm	$	8.88	(3.0)	&	\bfseries 42.96	$	\pm	$	6.33	(1.0)	&	63.54	$	\pm	$	3.78	(4.0) \\
Voice	&	16.07	$	\pm	$	1.81	(3.0)	&	\bfseries 15.69	$	\pm	$	1.98	(1.0)	&	15.85	$	\pm	$	1.61	(2.0)	&	81.44	$	\pm	$	2.32	(4.0) \\
Song	&	\bfseries 57.57	$	\pm	$	4.27	(1.0)	&	60.38	$	\pm	$	4.93	(3.0)	&	58.98	$	\pm	$	5.67	(2.0)	&	94.26	$	\pm	$	2.09	(4.0) \\
Adult	&	75.67	$	\pm	$	0.99	(3.0)	&	72.91	$	\pm	$	1.33	(2.0)	&	\bfseries 71.76	$	\pm	$	0.84	(1.0)	&	86.40	$	\pm	$	0.95	(4.0) \\
Default	&	\bfseries 81.23	$	\pm	$	0.61	(1.0)	&	82.43	$	\pm	$	0.79	(3.0)	&	82.31	$	\pm	$	1.00	(2.0)	&	94.01	$	\pm	$	0.33	(4.0) \\

Flickr	&	70.93	$	\pm	$	1.03	(3.0)	&	67.74	$	\pm	$	1.33	(2.0)	&	\bfseries 67.40	$	\pm	$	1.14	(1.0)	&	95.97	$	\pm	$	0.54	(4.0) \\
Fera	& \bfseries	80.15	$	\pm	$	0.92	(1.0)	&	80.72	$	\pm	$	1.09	(2.0)	&	80.86	$	\pm	$	1.03	(3.0)	&	83.56	$	\pm	$	2.56	(4.0) \\

\hline

WQplants	& \bfseries	90.75	$	\pm	$	2.86	(1.0)	&	90.85	$	\pm	$	2.27	(2.0)	&	91.79	$	\pm	$	2.83	(3.0)	&	92.36	$	\pm	$	3.88	(4.0) \\
WQanimals	&	95.19	$	\pm	$	2.62	(2.0)	&	95.47	$	\pm	$	2.22	(3.0)	&	\bfseries 95.09	$	\pm	$	3.01	(1.0)	&	97.83	$	\pm	$	2.15	(4.0) \\
Thyroid	&	\bfseries 21.63	$	\pm	$	0.94	(1.0)	&	22.76	$	\pm	$	1.32	(2.0)	&	23.15	$	\pm	$	1.41	(3.0)	&	25.29	$	\pm	$	1.16	(4.0) \\

\hline

Rf1	&	\bfseries 47.51	$	\pm	$	2.11	(1.0)	&	70.88	$	\pm	$	1.69	(2.0)	&	72.36	$	\pm	$	1.92	(3.0)	&	93.58	$	\pm	$	0.81	(4.0) \\
Pain	& \bfseries	24.07	$	\pm	$	1.50	(1.0)	&	24.74	$	\pm	$	1.41	(2.0)	&	24.76	$	\pm	$	1.40	(3.0)	&	24.86	$	\pm	$	1.55	(4.0) \\
Disfa	& \bfseries	60.24	$	\pm	$	1.42	(1.0)	&	60.96	$	\pm	$	1.15	(3.0)	&	60.51	$	\pm	$	1.50	(2.0)	&	63.40	$	\pm	$	1.62	(4.0) \\
WaterQuality	&	99.43	$	\pm	$	0.46	(2.0)	&	\bfseries 99.06	$	\pm	$	0.60	(1.0)	&	99.72	$	\pm	$	0.43	(3.5)	&	99.72	$	\pm	$	0.43	(3.5) \\
Oes97	& \bfseries	95.24	$	\pm	$	4.02	(1.0)	&	95.84	$	\pm	$	3.53	(2.0)	&	97.03	$	\pm	$	3.73	(3.0)	&	100	$	\pm	$	0.00	(4.0) \\
Oes10	&	90.58	$	\pm	$	3.45	(2.0)	&	\bfseries 90.33	$	\pm	$	4.62	(1.0)	&	92.30	$	\pm	$	3.26	(3.0)	&	100	$	\pm	$	0.00	(4.0) \\
Scm20d	& \bfseries	87.79	$	\pm	$	1.25	(1.0)	&	95.46	$	\pm	$	0.96	(4.0)	&	93.89	$	\pm	$	1.29	(3.0)	&	92.58	$	\pm	$	0.98	(2.0) \\
Scm1d	& \bfseries	80.75	$	\pm	$	1.02	(1.0)	&	89.86	$	\pm	$	2.17	(3.0)	&	88.81	$	\pm	$	0.88	(2.0)	&	90.81	$	\pm	$	1.01	(4.0) \\

\hline
Ave. rank		&			\bfseries			1.55		&							2.20		&							2.38		&							3.88	\\
    \bottomrule
    \end{tabular}
    \label{tab:lr_ss}
\end{table}

\begin{table}
    \centering
    \caption{Subset 0/1 loss (mean $\pm$ std.) of each MDC approach (\textbf{base learner: \textit{Naive Bayes}}).}
    \label{tab:nb_ss}
    \begin{tabular}{p{5em}|p{7.15em}p{7.45em}p{7.65em}p{7.15em}}
    \toprule
    \multirow{2}{*}{Data Set} & \multicolumn{4}{c}{Subset 0/1 loss (in \%)} \\ \cmidrule(r){2-5}
    & \multicolumn{1}{c}{GBNC-S} & BR & CC & CP  \\ \hline
Edm	& \bfseries 48.79	$	\pm	$	8.25	(1.0)	&	 57.12	$	\pm	$	4.34	(4.0)	& 52.13	$	\pm	$	9.22	(2.0)	&	\small 55.08	$	\pm	$	15.36	(3.0) \\
Jura	&	65.44	$	\pm	$	7.07	(2.0)	&	69.33	$	\pm	$	5.45	(3.0)	& \bfseries	64.05	$	\pm	$	6.26	(1.0)	&	98.89	$	\pm	$	1.36	(4.0) \\
Enb	& \bfseries	45.20	$	\pm	$	4.43	(1.0)	&	59.25	$	\pm	$	5.62	(3.0)	&	58.99	$	\pm	$	6.36	(2.0)	&	62.51	$	\pm	$	5.81	(4.0) \\
Voice	&	\bfseries 17.92	$	\pm	$	1.23	(1.0)	&	21.62	$	\pm	$	2.46	(2.0)	&	22.71	$	\pm	$	2.94	(3.0)	&	84.82	$	\pm	$	2.11	(4.0) \\
Song	&	\bfseries 70.72	$	\pm	$	4.57	(1.0)	&	78.60	$	\pm	$	4.13	(3.0)	&	77.71	$	\pm	$	3.23	(2.0)	&	93.63	$	\pm	$	3.26	(4.0) \\
Adult	&	92.75	$	\pm	$	0.68	(3.0)	&	76.69	$	\pm	$	3.14	(2.0)	& \bfseries	74.65	$	\pm	$	2.80	(1.0)	&	98.46	$	\pm	$	0.35	(4.0) \\
Default	& \bfseries	92.11	$	\pm	$	1.89	(1.0)	&	96.10	$	\pm	$	2.23	(2.0)	&	96.47	$	\pm	$	1.91	(3.0)	&	99.94	$	\pm	$	0.03	(4.0) \\
Flickr	& \bfseries	82.94	$	\pm	$	1.33	(1.0)	&	86.03	$	\pm	$	1.11	(3.0)	&	85.89	$	\pm	$	1.08	(2.0)	&	96.56	$	\pm	$	0.55	(4.0) \\
Fera	&	\bfseries 93.34	$	\pm	$	0.37	(1.0)	&	97.94	$	\pm	$	0.35	(3.0)	&	98.04	$	\pm	$	0.31	(4.0)	&	95.69	$	\pm	$	0.53	(2.0) \\

\hline

WQplants	& \bfseries	99.15	$	\pm	$	0.66	(1.5)	&	100	$	\pm	$	0.00	(4.0)	&	99.91	$	\pm	$	0.28	(3.0)	&	\bfseries 99.15	$	\pm	$	0.89	(1.5) \\
WQanimals	& \bfseries	99.15	$	\pm	$	0.66	(1.0)	&	99.62	$	\pm	$	0.86	(3.5)	&	99.43	$	\pm	$	0.96	(2.0)	&	99.62	$	\pm	$	0.63	(3.5) \\
Thyroid	& \bfseries	20.39	$	\pm	$	1.92	(1.0)	&	90.96	$	\pm	$	0.95	(4.0)	&	88.79	$	\pm	$	1.15	(3.0)	&	47.24	$	\pm	$	2.33	(2.0)\\

\hline

Rf1	&	\bfseries 69.12	$	\pm	$	1.09	(1.0)	&	83.82	$	\pm	$	0.97	(2.5)	&	83.82	$	\pm	$	0.65	(2.5)	&	93.05	$	\pm	$	0.60	(4.0) \\
Pain	&	89.14	$	\pm	$	0.77	(2.0)	&	93.08	$	\pm	$	0.80	(3.0)	& \bfseries	87.49	$	\pm	$	1.38	(1.0)	&	91.56	$	\pm	$	1.10	(4.0) \\
Disfa	& \bfseries	88.62	$	\pm	$	2.03	(1.0)	&	99.66	$	\pm	$	0.14	(4.0)	&	99.47	$	\pm	$	0.17	(3.0)	&	94.58	$	\pm	$	1.02	(2.0) \\
WaterQuality	&	100	$	\pm	$	0.00	(3.0)	&	100	$	\pm	$	0.00	(3.0)	&	100	$	\pm	$	0.00	(3.0)	& \bfseries	99.72	$	\pm	$	0.43	(1.0) \\
Oes97	&	96.69	$	\pm	$	3.12	(3.0)	& \bfseries	94.30	$	\pm	$	4.50	(1.0)	&	94.60	$	\pm	$	4.36	(2.0)	&	100	$	\pm	$	0.00	(4.0) \\
Oes10	&	91.79	$	\pm	$	3.91	(2.0)	& \bfseries	91.54	$	\pm	$	4.09	(1.0)	&	92.53	$	\pm	$	4.21	(3.0)	&	99.01	$	\pm	$	1.22	(4.0) \\
Scm20d	&	98.63	$	\pm	$	0.32	(4.0)	&	98.27	$	\pm	$	0.30	(3.0)	&	97.60	$	\pm	$	0.43	(2.0)	&	\bfseries 96.07	$	\pm	$	0.43	(1.0) \\
Scm1d	&	95.54	$	\pm	$	1.25	(3.0)	&	91.66	$	\pm	$	0.70	(2.0)	&	\bfseries 91.03	$	\pm	$	0.91	(1.0)	&	96.59	$	\pm	$	0.58	(4.0) \\
\hline
Ave. rank	&			\bfseries				1.73		&							2.80		&							2.28		&							3.20	\\
    \bottomrule
    \end{tabular}
\end{table}

\end{document}